\documentclass[preprint,12pt]{elsarticle}

\usepackage{times}
\usepackage{epsfig}
\usepackage{graphicx}
\usepackage{amsmath}
\usepackage{amssymb}
\usepackage{comment}
\usepackage{amsmath}
\usepackage{booktabs}
\usepackage{tabularx}
\usepackage{epsfig}
\usepackage{pifont}
\usepackage[dvipsnames]{xcolor}
\usepackage{hyperref}
\newcolumntype{M}[1]{>{\centering\arraybackslash}m{#1}}
\newcolumntype{C}[1]{>{\centering\let\newline\\\arraybackslash\hspace{0pt}}p{#1}}
\newcolumntype{R}[1]{>{\raggedleft\let\newline\\\arraybackslash\hspace{0pt}}p{#1}}
\newcolumntype{L}[1]{>{\raggedright\let\newline\\\arraybackslash\hspace{0pt}}p{#1}}

\newcommand\Tstrut{\rule{-3pt}{2.6ex}}       
\newcommand\Bstrut{\rule[-0.9ex]{-3pt}{0pt}} 
\usepackage{multirow}
\usepackage{adjustbox}
\DeclareMathAlphabet{\mymathbb}{U}{bbold}{m}{n}
\usepackage{dblfloatfix}
\usepackage{setspace} 
\usepackage{float}
\usepackage{lipsum}
\usepackage{caption}
\journal{Pattern Recognition}
\begin{document}

\begin{frontmatter}

\title{MKConv: Multidimensional Feature Representation for \\ Point Cloud Analysis}

\author{Sungmin Woo}
\ead{smw3250@yonsei.ac.kr}
\author{Dogyoon Lee}
\author{Sangwon Hwang}
\author{Woo Jin Kim}
\author{Sangyoun Lee\corref{mycorrespondingauthor}}
\ead{syleee@yonsei.ac.kr}

\cortext[mycorrespondingauthor]{Corresponding author}
\address{School of Electrical and Electronic Engineering, Yonsei University, \\ Seoul 03722, South Korea}

\begin{abstract}
Despite the remarkable success of deep learning, an optimal convolution operation on point clouds remains elusive owing to their irregular data structure. Existing methods mainly focus on designing an effective continuous kernel function that can handle an arbitrary point in continuous space. Various approaches exhibiting high performance have been proposed, but we observe that the standard pointwise feature is represented by 1D channels and can become more informative when its representation involves additional spatial feature dimensions. In this paper, we present Multidimensional Kernel Convolution (MKConv), a novel convolution operator that learns to transform the point feature representation from a vector to a multidimensional matrix. Unlike standard point convolution, MKConv proceeds via two steps. (i) It first activates the spatial dimensions of local feature representation by exploiting multidimensional kernel weights. These spatially expanded features can represent their embedded information through spatial correlation as well as channel correlation in feature space, carrying more detailed local structure information. (ii) Then, discrete convolutions are applied to the multidimensional features which can be regarded as a grid-structured matrix. In this way, we can utilize the discrete convolutions for point cloud data without voxelization that suffers from information loss. Furthermore, we propose a spatial attention module, Multidimensional Local Attention (MLA), to provide comprehensive structure awareness within the local point set by reweighting the spatial feature dimensions. We demonstrate that MKConv has excellent applicability to point cloud processing tasks including object classification, object part segmentation, and scene semantic segmentation with superior results.
\end{abstract}

\end{frontmatter}
\section{Introduction}
Point cloud analysis has attained increasing significance in recent years as 3D vision, which relies on point cloud data, has become an essential requirement in an extensive range of applications such as autonomous driving, robotics, and augmented reality. However, it is not straightforward to process point clouds because of their unique properties of being sparse, unordered, and irregular. This non grid-structured data cannot be directly handled by superior 2D deep learning methods designed for grid structured data. 

One simple approach is to convert a point cloud into 3D voxel grids~\cite{wu20153d,tchapmi2017segcloud,graham20183d}. Standard discrete convolution then can be performed on discrete voxels by encoding the geometric attributes of points contained therein. However, voxelization of the raw point cloud produces a vast number of voxels~\citep{qi2017pointnet}, and 3D convolutions on such data representation involve high complexity and inefficient memory consumption that increase with the resolution. Furthermore, it inherently fails to fully use the given data because information loss is inevitable in quantization~\citep{qi2017pointnet,liu2019relation,jiang2018pointsift}. 

Several studies have overcome these limitations by applying convolution directly to the points (i.e., point convolution). A key challenge for point convolution is that kernel weights for an arbitrary position within the continuous receptive field should be obtainable to cope with the irregularity of the point cloud. Therefore, early works~\citep{hermosilla2018monte,wu2019pointconv} constructed spatially continuous kernels by training Multi-Layer Perceptrons (MLPs) as a kernel function to directly output the weights with the relative position as an input. However, given that producing large kernel weights using MLPs is computationally burdensome, recent works subdivided the process of constructing continuous kernels. \cite{boulch2019generalizing,boulch2020convpoint,thomas2019kpconv} devised kernel points whose weights were parameterized and used interpolation to define the entire kernel from kernel points. \cite{liu2019relation} used MLPs to learn a small size of channel-wise kernel weights and raised channels with additional MLPs. \cite{xu2021paconv} defined a weight bank wherein several kernel weights were parameterized and used MLP to assemble the kernel weights.

\begin{figure*}[t]
	\begin{center}
		\includegraphics[width=1\linewidth]{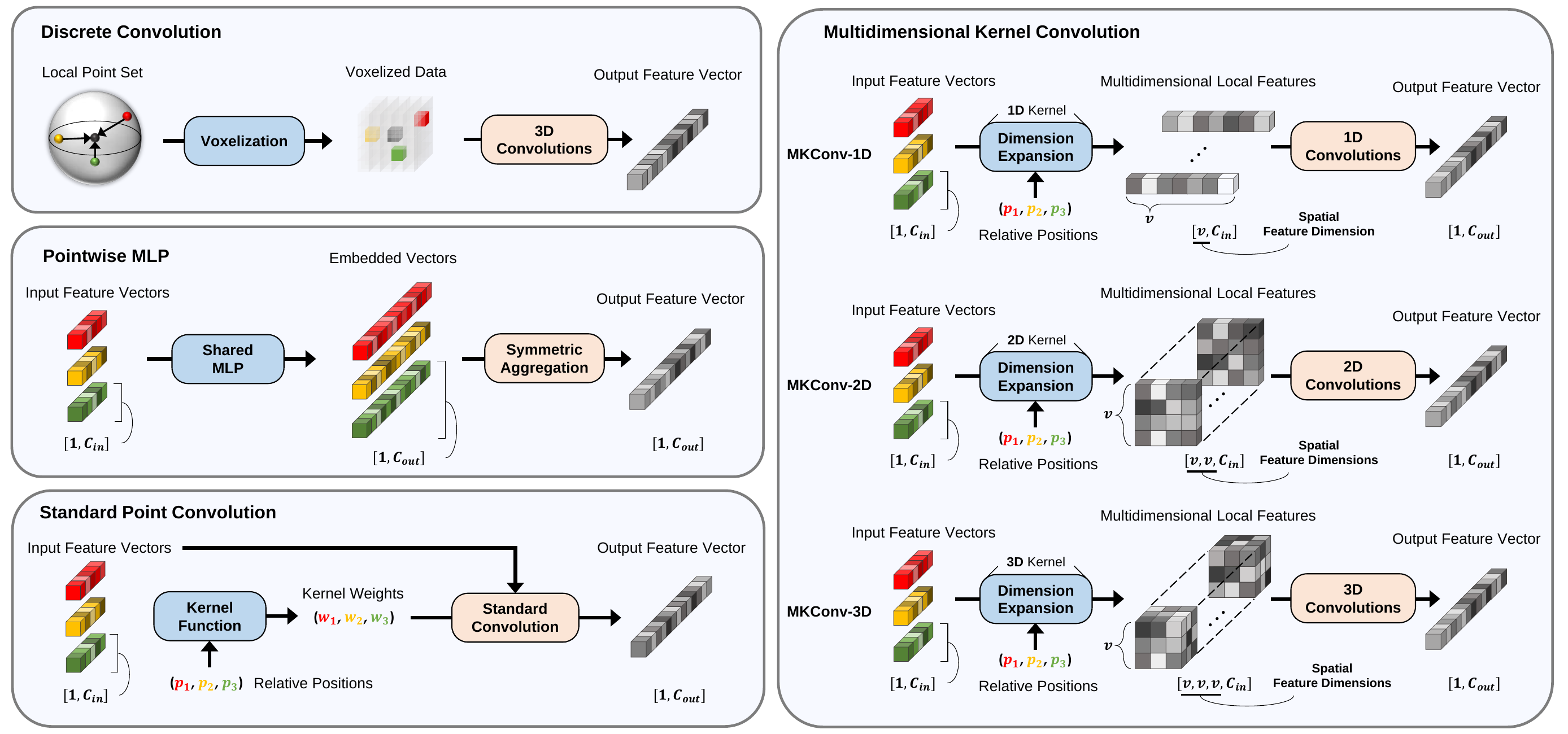}
		
	\end{center}
	\vspace{-0.5cm}
	\caption{Overview of existing methods (left) and MKConv (right). MKConv can carry detailed geometric information via additional spatial arrangements of multidimensional local features, unlike other methods in which features have no spatial dimension.
	}
	
	\label{fig:1p}
\end{figure*}

Although these existing methods for point clouds exhibit high performance, we observe that their feature representations are limited to the channel dimension as in Figure~\ref{fig:1p}. Similar to pixel-wise feature representation in image data, each point represents its embedded information with only one dimension of a  feature vector, i.e., channels. This motivates us to further explore the spatial dimensions in feature space, seeking a better form of message passing. Accordingly, we propose a novel point convolution operator named Multidimensional Kernel Convolution (MKConv) that learns to activate the spatial dimensions of feature representation. 

\begin{figure}[t]
	\begin{center}
		\includegraphics[width=1\linewidth]{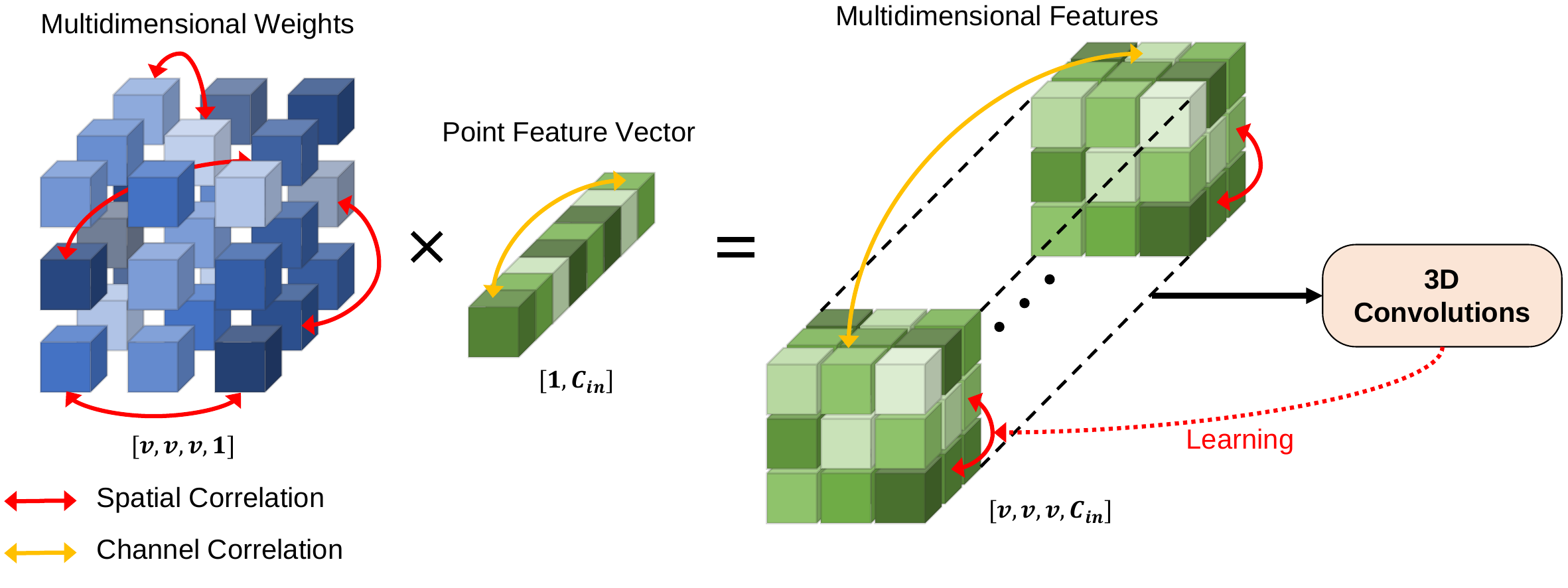}
	\end{center}
	\vspace{-0.7cm}
	\caption{Spatial dimension expansion for a point feature vector when kernel weight dimension is 3D. Unlike standard point convolution, MKConv learns multidimensional feature-level spatial correlation between weights that determines the geometrical arrangements of multidimensional features in embedding space. The spatial correlation is learned by the feedback from following convolutions, which operate spatially on multidimensional features.
	}
	
	\label{fig:2p}
\end{figure}

Specifically, MKConv adopts multidimensional kernel weights that are obtained by learning the continuous kernel function, which extracts the feature-level spatial correlation between weights. As depicted in Figure~\ref{fig:2p}, the learned spatial correlation among multidimensional weights determines the spatial arrangements of multidimensional features in the embedding space, enabling the spatial expansion of feature representation. The expanded features of neighboring points in the local point set are aggregated and then processed by discrete convolutions. Owing to their additional representation ability in spatial dimensions, these multidimensional features can reflect the detailed structure of local points more concretely. In addition, unlike other point convolution methods, MKConv can exploit discrete convolutions without the information loss that inevitably ensues during quantization. Our dimension activation in feature space rather enriches the information while producing a grid-structured matrix concurrently.

Furthermore, we propose Multidimensional Local Attention (MLA) which provides additional spatial attention to the multidimensional features with comprehensive structure awareness of the local point set. The learnable MLA reweights the spatial arrangement of multidimensional features to emphasize the informative part in spatial feature dimensions. Through local-structure-aware guidance provided by MLA, multidimensional features can be more representative. 

We evaluate the proposed approach on three tasks: object classification, object part segmentation, and scene semantic segmentation. The experimental results demonstrate the effectiveness and superiority of MKConv, with state-of-the-art results among point convolution methods.

Our main contributions can be summarized as follows:

\begin{itemize}
	\setlength{\itemsep}{0ex}
	\item To the best of our knowledge, this is the first study on learning to activate the multiple spatial dimensions in feature space for better feature representation;
	\item We propose a novel MKConv for effective point cloud analysis. The dimension expansion of MKConv enables (i) spatially correlated feature representation and (ii) discrete convolutions without information loss;
	\item We introduce learnable MLA, which reweights the multidimensional features using comprehensive structural information;
	\item We conduct extensive experiments on various point cloud processing tasks with theoretical analysis while achieving superior performance.
\end{itemize}


\section{Related Works}
\label{related}
\noindent\textbf{Discrete convolution methods.}\hspace{0.3cm}Early works exploited regular grid convolution by converting a point cloud into a grid representation, such as 2D pixels or 3D voxels. For 2D representation, a point cloud can be transformed into multi-view images~\citep{yang2019learning, liu2022vfmvac} or a range image~\citep{wu2019squeezesegv2,milioto2019rangenet++} and fast 2D convolutions are applied. For 3D representation, the 3D space is discretized into a set of occupancy voxels \citep{wu20153d,tchapmi2017segcloud}. As volumetric data has empty space where no value is assigned, some studies \cite{graham20183d,su2018splatnet} propose methods to learn from occupied voxels, efficiently reducing memory and computational cost. However, inevitably, the discretization of non-grid structured data results in the loss of detailed geometric information. In contrast, our approach does not suffer from information loss because it takes a raw point cloud without transforming the data representation. Instead, MKConv learns to transform the dimension of feature representation in embedding space, enabling discrete convolutions on the spatially expanded features.

\medskip	
\noindent\textbf{Pointwise MLP methods.}\hspace{0.3cm}As a pioneering work, PointNet~\citep{qi2017pointnet} proposes to exploit shared MLPs and the symmetric aggregation function to process the point cloud directly. Pointwise features are learned independently from an MLP that is shared over points, and global features are extracted by a max-pooling operation to achieve permutation invariance. PointNet++~\citep{qi2017pointnet++} designs a hierarchical network to further capture neighborhood information for each point. Local geometric features are learned by applying PointNet to local groups of points. 
Furthermore, several other methods~\citep{wang2019dynamic,liu2019densepoint,li2018so, zhang2019shellnet} basically adopt the MLP shared for grouped points and aggregate the features with diverse functions. However, shared MLPs that directly transform point features are limited to capturing spatial-variant information, as mentioned by \cite{xu2021paconv}.

\medskip	
\noindent\textbf{Transformer methods.}\hspace{0.3cm}Recently, various methods~\cite{guo2021pct, zhang2022pvt, zhang2022patchformer} adopt transformer architecture to process the point cloud data, taking advantage of its flexible and strong feature extraction ability. PCT~\cite{guo2021pct} proposes the offset-attention layer that calculates the offset between self-attention features and input features by element-wise subtraction. 
PVT~\cite{zhang2022pvt} combines the idea of voxel-based and point-based networks into transformer architecture to efficiently encode coarse-grained features and effectively aggregate fine-grained features.

\medskip	
\noindent\textbf{Point convolution methods.}\hspace{0.3cm}Point convolution methods tend to extend the image convolution concept. These methods define a continuous kernel function to obtain pointwise weights.
PointConv~\citep{wu2019pointconv} exploits the outer product and $1 \times 1$ convolution to reduce the memory consumption required in computing kernel weights. Here, the feature matrix obtained from the outer product is flattened to a feature vector and arranged in a channel dimension. In contrast, the outer product used in our MKConv produces a multidimensional matrix which involves the spatial correlation. MKConv activates the spatial feature dimensions by learning multidimensional weights and discrete convolutions. KPConv~\citep{thomas2019kpconv} defines kernel points that carry kernel weights. The weights for kernel points can be directly parameterized, but the interpolation function that approximates the entire continuous kernel from kernel points is manually defined. In contrast, the entire kernel function of MKConv is learnable, enhancing the optimization capacity. FPConv~\citep{lin2020fpconv} learns a weight map to project points onto a 2D grid and applies 2D convolution on the image form of features. This idea is similar to that of MKConv, given that it discretizes a local point set in feature space. However, FPConv has the severe limitation that its 2D output features lack the ability to represent object curvature. In contrast, MKConv can adopt diverse spatial feature dimensions including 3D, which provide an expanded feature space to represent embedded information. PAConv~\citep{xu2021paconv} defines a weight bank that contains several parameterized kernel weights, and MLP is used to assemble kernel weights by learning from the relation between points. Unlike existing point convolution methods that focus on designing a continuous kernel function, we explore the spatial dimensions of feature representation in feature space for better description ability in terms of local geometry.

\section{Multidimensional Kernel Convolution}
We first briefly introduce the principle of the point convolution operation and then expound on our proposed Multidimensional Kernel Convolution (MKConv). 

\subsection{Preliminaries}

\noindent\textbf{Notations.}\hspace{0.3cm}For the sake of clarity, we define the notations employed in the paper as follows. 

Neighboring points around a point $p$ within a predefined radius $r \in \mathbb{R}$ are denoted as 
$\mathcal{N}(p)=\left\lbrace q \mid \left\Vert q-p \right\Vert < r \right\rbrace$, where $p,q \in \mathbb{R}^3$ are cartesian coordinates. For MKConv, we randomly sample a set of $N$ neighboring points $\mathcal{N}_s(p)$ from $\mathcal{N}(p)$ in order to ensure robustness to point cloud density, because the number of points in $\mathcal{N}(p)$ differs depending on the query point $p$. The points in $\mathcal{N}_s(p) \in \mathbb{R}^{N\times3}$ are regarded as a \textit{local point set}, where a single convolution operation is performed. The feature derived at a point $q$ is denoted as $f(q) \in \mathbb{R}^{C_{in}}$, where $C_{in}$ is the number of input channels, and $f(q)$ can be initialized using additional information such as the normal vector or RGB color. We denote the standard convolution kernel function and our proposed multidimensional kernel function as $g_s$ and $g_m$, respectively. Both convolution kernel functions determine kernel weights in continuous receptive fields.

\medskip
\noindent\textbf{Standard point convolution formulation.}\hspace{0.3cm}As shown in previous works~\citep{hermosilla2018monte,wu2019pointconv,thomas2019kpconv,xu2021paconv}, the standard convolution operation on an arbitrary point $p$ can be formulated as
\begin{equation} \label{eq:1}
	(f*g_s)(p) = \sum_{q_i \in \mathcal{N}(p)} \hspace{-0.2cm} f(q_i)^Tg_s(q_i-p),
\end{equation}
where $g_s : \mathbb{R}^3 \to \mathbb{R}^{C_{in} \times C_{out}}$ produces weights for the neighboring point $q_i$,  with output channel size $C_{out}$.
This formulation is essentially identical to image convolution except for the characteristic of the kernel function $g_s$. As a point cloud is non grid-structured data without fixed positions, $g_s$ should be able to handle any point in the continuous kernel space. Thus, the kernel function must be designed to obtain point-dependent kernel weights based on the positions of points. In general, the weights for $q_i$ are learned from its relative position $q_i-p$.

\subsection{Observation}
In standard point convolution, the feature derived at an arbitrary point $p$ is $f(p) \in \mathbb{R}^{C}$, which is a vector comprised of channels having size $C$. Specifically, each channel of a point feature vector is a non-dimensional value, i.e., \textit{scalar feature}. Existing point convolution methods adopt this scalar feature representation where the embedded information is represented by 1D arrangement of scalar features (correlation between channels). In contrast, we propose to further explore the potential of feature representation in the spatial dimensions where feature can achieve better capacity of expression by feature-level spatial correlation. The key insight in this work is that feature representation with spatial correlation as well as channel correlation can better deliver the embedded information than that with channel correlation only. To this end, we design a learnable spatial dimension expansion by educing the interaction between continuous multidimensional kernel function and discrete convolutions. Our proposed MKConv learns to produce the multidimensional feature representation, activating the spatial feature dimensions and enriching embedded information carried throughout point convolution.

\subsection{Multidimensional Feature Representation}
\label{sec:fv}
We propose the dimension expansion process, which activates the spatial dimensions of local feature representation. The key idea is to adopt the multidimensional kernel weights that map features from the 1D vector (one channel dimension) to the multidimensional matrix (one channel dimension and additional spatial dimensions).
We define the k-dimensional kernel weights $w_k$ for a neighboring point $q$ with a query point $p$ from the k-dimensional kernel function $g_{m,k}$ as
\begin{equation} \label{eq:2}
	w_k^q = g_{m,k}(q-p) = \mathcal{R}_{1D \to kD}(\theta_k(q-p)) \in \mathbb{R}^{\begin{scriptsize}\overbrace{v_k \times \cdots \times v_k}^k\end{scriptsize}},
\end{equation}
where $\theta_k : \mathbb{R}^3 \to \mathbb{R}^{({v_k})^k}$ is a non-linear function implemented with shared MLPs to be invariant to the input order of points as in certain previous works~\citep{wu2019pointconv,liu2019relation,lin2020fpconv}. $\mathcal{R}_{1D \to kD} : \mathbb{R}^{({v_k})^k} \to \mathbb{R}^{\begin{scriptsize}\overbrace{v_k \times \cdots \times v_k}^k\end{scriptsize}}$ is the reshaping function, and $v_k$ is the predefined k-dimensional kernel unit size that determines the spatial resolution of multidimensional features. With the k-dimensional kernel weights $w_k^q$, the point feature vector $f(q) \in \mathbb{R}^{C_{in}}$ can be spatially expanded to the multidimensional point feature matrix $f_{m,k}(q) \in \mathbb{R}^{\begin{scriptsize}\overbrace{v_k \times \cdots \times v_k}^k \times C_{in}\end{scriptsize}}$ as follows:\begin{figure*}[t]
	\begin{center}
		\includegraphics[width=1\linewidth]{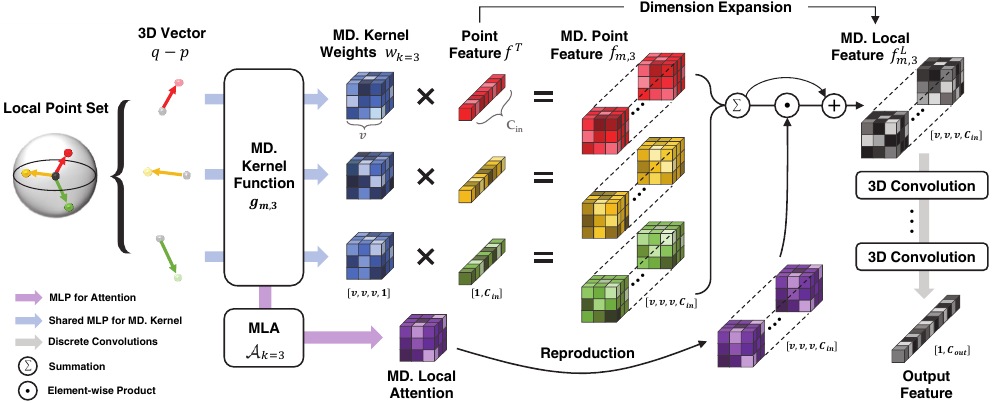}	
	\end{center}
	\vspace{-0.5cm}
	\caption{Overall MKConv-3D (k=3) process on a local point set with number of neighboring points $N=3$ and kernel unit size $v_{k=3} = 3$. Dimension expansion with the multidimensional kernels is performed using Eq.~\ref{eq:5}, and multidimensional attention is reproduced on the channel axis for an element-wise product with multidimensional local features. Detailed architectures of the multidimensional kernel function and MLA are depicted in Figure~\ref{fig:attention}.
	}
	
	\label{fig:main}
\end{figure*}
\begin{equation} \label{eq:3}
	f_{m,k}(q) = w_k^q \otimes f(q),
\end{equation}
where $\otimes$ is the outer product operation.
In other words, each scalar feature (channel) of $f(q)$ is independently multiplied by the multidimensional weights $w_k^q$ to be expanded to $f_{m,k}(q)$. As it aims to learn the spatial expansion for a single scalar feature, the correlation between channels is excluded by Eq.~\ref{eq:3}.
The multidimensional local features $f_{m,k}^L \in \mathbb{R}^{\begin{scriptsize}\overbrace{v_k \times \cdots \times v_k}^k \times C_{in}\end{scriptsize}}$ for the query point $p$ are then obtained by aggregating the multidimensional point features of neighboring points:
\begin{equation} \label{eq:5}
	\begin{gathered}
		f_{m,k}^L(p) = \hspace{-0.1cm}\sum_{q_i \in \mathcal{N}_s(p)} \hspace{-0.2cm}f_{m,k}(q_i) = \hspace{-0.1cm}\sum_{q_i \in \mathcal{N}_s(p)} \hspace{-0.2cm} (g_{m,k}(q_i-p) \otimes f(q_i)).
	\end{gathered}
\end{equation}
To facilitate understanding, we illustrate the overall MKConv process with k=3 in Figure~\ref{fig:main}.

While our dimension expansion produces multidimensional features from the 1D feature vector, it requires only a small number of parameters. The size of the multidimensional kernel weights is fixed at $(v_k)^k$, which is independent of the feature channel size $C_{in}$, and the predefined $v_k$ need not be large as discussed in Section~\ref{sec:ablation}. We set $v_{k=3}=4$ for MKConv-3D based on the experimental results presented in Table~\ref{tab:v}.

\medskip
\noindent\textbf{Enforcing dimension expansion learning.}\hspace{0.3cm}When we only observe the process of dimension expansion, the arrangement of spatial feature dimensions of $f^L_{m,k}$ is not spatially correlated as multidimensional kernel weights are simply obtained by reshaping the dimension of 1D weight vector in the function $\mathcal{R}$ : $\mathbb{R}^{(v_k)^k} \to \mathbb{R}^{\begin{scriptsize}\overbrace{v_k \times \cdots \times v_k}^k\end{scriptsize}}$ (Eq.~\ref{eq:2}). However, we can force the learning of multidimensional kernel function $g_{m,k}$ to produce k-dimensionally ordered kernel weights by applying k-dimensional convolutions on the multidimensional local features $f^L_{m,k}$. For example, consider MKConv-3D shown in Figure~\ref{fig:main} where $k=3$. As $g_{m,3}$ is learnable and the following 3D convolutions require $f^L_{m,3}$ to have proper 3D spatial arrangements, $g_{m,k}$ learns to produce the spatially-aware multidimensional weights that map features from scalar to 3D during training. This interaction between dimension expansion process and discrete convolutions is enforced because they are connected in series, affecting each other via gradient backpropagation. In other words, the spatial feature dimension mapping in $g_{m,3}$ is gradually learned through the feedback from 3D convolutions, which operate spatially on $f^L_{m,3}$ to extract the correlation along spatial feature dimensions. We validate the influence of learning spatially expanded multidimensional feature representation in Section~\ref{sec:ablation}.

Here, it is important to note that the spatial activation of $f^L_{m,k}$ in feature space does not correspond to the local point distribution of $\mathcal{N}_s$ in physical space. Interacting with the following k-dimensional convolutions, $g_{m,k}$ learns to derive a spatial correlation in feature space (see Figure~\ref{fig:2p}), instead of imitating the physical distribution of points. Similar to feature-level information within the channel dimension, the expanded spatial feature dimensions offer another way to represent the embedded information.

\subsection{Multidimensional Kernel Weight Normalization}
As a multidimensional kernel is obtained from the continuous function $g_m$ approximated by MLPs, the scale and variance of the multidimensional weights vary substantially from point to point. We prevent this weight imbalance from causing unstable training by imposing restrictions on the distribution of multidimensional weights with normalization schemes. 

Let the vector form of multidimensional weights $w_k$ be $\hat{w}_k=\left\lbrace \hat{w}_{k,i} \mid i=1,2, ..., (v_k)^k \right\rbrace \in \mathbb{R}^{(v_k)^k}$ for a given point. The L2 normalization on each weight to stabilize scale distributions of multidimensional weights over points can be formulated as
\begin{equation} \label{eq:6}
	\textsc{Norm}_{\textsc{L2}}(\hat{w}_{k,i}) =  \frac{\hat{w}_{k,i}}{\sqrt{\sum \limits _{j=1} ^{(v_k)^k} (\hat{w}_{k,j})^2}}.
\end{equation}
Standardization can be alternatively applied to further restrict both scale and variance distributions of multidimensional weights, as follows:
\begin{equation} \label{eq:7}
	\textsc{Norm}_{\textsc{st}}(\hat{w}_{k,i}) =  \frac{\hat{w}_{k,i}-\mu}{\sigma},
\end{equation}
where $\mu = \cfrac{1}{(v_k)^k} \sum \limits _{j=1} ^{(v_k)^k} \hat{w}_{k,j}$ and $\sigma = \sqrt{\cfrac{1}{(v_k)^k} \sum \limits _{j=1} ^{(v_k)^k} (\hat{w}_{k,j} - \mu)^2}$. These multidimensional weight normalization schemes help to improve performance (Section~\ref{sec:ablation}), and hence we apply normalization after every weight prediction in the MKConv layers.

\begin{figure}[t]
	\begin{center}
		\includegraphics[width=0.8\linewidth]{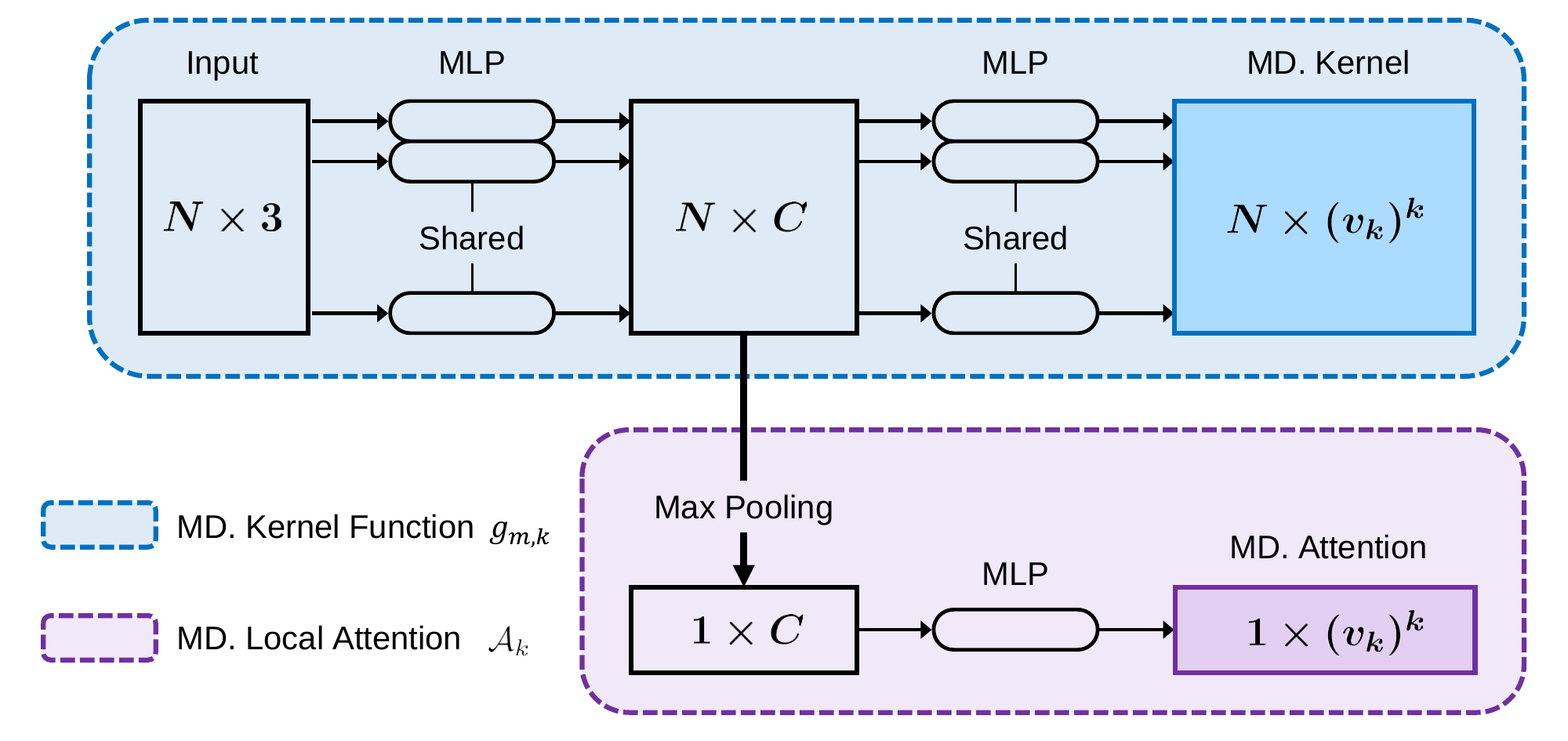}
	\end{center}
	\vspace{-0.6cm}
	\caption{Multidimensional kernel function (top) and MLA (bottom) architectures. Note that both outputs are reshaped to the multidimensional matrix afterwards (i.e., $\mathbb{R}^{(v_k)^k} \to \mathbb{R}^{\begin{scriptsize}\overbrace{v_k \times \cdots \times v_k}^k\end{scriptsize}}$).
	}
	
	\label{fig:attention}
\end{figure}

\subsection{Multidimensional Local Attention}
Given the irregular nature of point clouds, every local point set has a unique structure. Thus, it is critically important that point convolution captures the distinct structures of respective local point sets. However, different from standard point convolution which only aggregates the features from neighboring points, MKConv needs to learn spatial dimension expansion for each point vector as well as feature aggregation. This dimension expansion learning requires an additional optimization of the multidimensional local features $f^L_m$ to reflect the comprehensive local structure information in both channel and spatial feature dimensions. Thus, we apply auxiliary attention that emphasizes the informative part in spatial feature dimensions based on the comprehensive local geometric information extracted from the local point set, helping the optimization of spatial feature dimensions and making the speed of convergence faster. Our learnable MLA reweights the spatial arrangement of $f^L_m$ by considering neighboring points in the local point set jointly. As spatial dimensions of MLA should be collinear in the feature space to those of $f^L_m$, the attention weights are obtained by sharing the multidimensional kernel function $g_m$ as depicted in Figure~\ref{fig:attention}. Unlike conventional attention mechanisms that multiply input features by the obtained attention scores again, MLA computes the spatial attention from the local distribution information and multiplies it by multidimensional local features in the manner of a skip connection (Figure~\ref{fig:main}), explicitly reweighting the spatial arrangement of $f^L_m$. This multidimensional spatial attention is specialized for MKConv, which adopts the multidimensional feature representation involving spatial dimensions.

Specifically, MLA shares the front MLP of the multidimensional kernel function to perform identical dimension expansion as shown in Figure~\ref{fig:attention}. The representative feature of the local point set is obtained by applying max pooling on the intermediate features, and another MLP is used to extract multidimensional attention. The multidimensional attention is then reproduced along the channel dimension for element-wise multiplication with multidimensional local features (i.e., $\mathbb{R}^{\begin{scriptsize}\overbrace{v_k \times \cdots \times v_k}^k\end{scriptsize}} \to \mathbb{R}^{\begin{scriptsize}\overbrace{v_k \times \cdots \times v_k}^k \times C_{in}\end{scriptsize}}$). 

Let the relative positions $\mathcal{H}$ of neighboring points with respect to a query point $p$, which is used as an input in Figure~\ref{fig:attention} be 
\begin{equation}
	\label{eq:8}
	\mathcal{H}(p)=\left\lbrace h_i \mid h_i = q_i - p, q_i \in \mathcal{N}_s(p) \right\rbrace \in \mathbb{R}^{N \times 3}.
\end{equation}
Then the multidimensional local features $f^L_{m,k}$ with MLA can be extended from Eq.~\ref{eq:5} as
\begin{equation}
	\label{eq:9}
	f^L_{m,k}(p) = (\mymathbb{1} + \mathcal{A}_k(\mathcal{H}(p))) \hspace{0.05cm} \odot \hspace{-0.2cm} \sum_{h_i \in \mathcal{H}(p)}\hspace{-0.2cm} (g_{m,k}(h_i) \otimes f(p+h_i)),
\end{equation}
where $\mathcal{A}_k : \mathbb{R}^{N \times 3} \to \mathbb{R}^{\begin{scriptsize}\overbrace{v_k \times \cdots \times v_k}^k \times C_{in}\end{scriptsize}}$ is the attention function producing multidimensional attention and $\odot$ is the element-wise product. $\mymathbb{1}$ is a matrix of ones having the same size as $\mathcal{A}_k(\mathcal{H}(p))$ and indicates a skip connection for features without MLA. Finally, our MKConv on the query point $p$ can be written as
\begin{equation}
	\label{eq:10}
	\text{MKConv}_k(p) = \text{Conv}_\text{k}(f^L_{m,k}(p)),
\end{equation}
where $\text{Conv}_\text{k} : \mathbb{R}^{\begin{scriptsize}\overbrace{v_k \times \cdots \times v_k}^k \times C_{in}\end{scriptsize}} \to \mathbb{R}^{1 \times C_{out}}$ indicates several k-dimensional convolution operations on multidimensional local features with output channel $C_{out}$.

\section{Experiments}
\label{exp}

We evaluate our proposed MKConv on three point cloud processing tasks: object classification, object part segmentation, and scene semantic segmentation. In this section, we analyze the performance of MKConv with benchmarking results and ablation studies. Unless otherwise stated, we adopt MKConv-3D ($k$=3) for our model and ablation studies on other feature dimensions $k$ are addressed in Section~\ref{sec:ablation}. Network architectures, configurations, detailed training parameters and more complexity analysis are provided in the supplementary material.

\subsection{Object Classification}
\label{sec:cls}

We use the ModelNet40~\citep{wu20153d} dataset, which contains 9,843 training and 2,468 test models in 40 classes for object classification. Point cloud data trained and tested with our model is provided by \cite{qi2017pointnet}, and 1,024 points are uniformly sampled as inputs. The points are normalized into a unit sphere. For training, random scaling and translation are used for the augmentation strategy as in \cite{klokov2017escape, liu2019relation}. A dropout layer with a probability of 0.5 is used in the final fully-connected (FC) layers to reduce the over-fitting problem.

\begin{table*}[t]
	\begin{center}
		\footnotesize
		\caption{Object classification results (\%) on ModelNet40. ``nr'', ``vot.'', ``Conv.'', ``Trans.'' denote normal vectors, voting strategy, convolution, and transformer, respectively.}
		\vspace{0cm}
		\begin{tabular}{l|cccc|c}
			\toprule
			Method & Reference & Operation & Input & \#points & OA  \\
			\midrule
			\midrule
			SpiderCNN \citep{xu2018spidercnn} & ECCV'18  & Conv.  &  xyz, nr & 1024   & 92.4  \\
			SO-Net \citep{li2018so}  & CVPR'18 & MLP   &  xyz, nr & 5000     & 93.4 \\
			PointConv~\citep{wu2019pointconv} & CVPR'19 & Conv.    &  xyz, nr & 1024    & 92.5  \\
			KPConv \citep{thomas2019kpconv}& ICCV'19 & Conv.  &  xyz & 6800   & 92.9 \\
			\midrule
			PointNet \citep{qi2017pointnet} & CVPR'17 & MLP   &  xyz & 1024  & 89.2 \\
			PointNet++ \citep{qi2017pointnet++}  & NIPS'17 & MLP  &  xyz & 1024 & 90.7 \\
			PointCNN \citep{li2018pointcnn} & NIPS'18 & Conv.   &  xyz & 1024 & 92.2   \\
			DGCNN~\citep{wang2019dynamic}  & TOG'19 & MLP  &  xyz & 1024  & 92.2\\
			RS-CNN~\citep{liu2019relation} w/o vot.& CVPR'19 & Conv. &  xyz &1024  & 92.9  \\
			RS-CNN~\citep{liu2019relation} w/ vot. & CVPR'19 & Conv. &  xyz &1024  & 93.6 \\
			InterpCNN~\citep{mao2019interpolated}  & ICCV'19 & Conv.  &  xyz& 1024   & 93.0 \\
			ShellNet~\cite{zhang2019shellnet} & ICCV'19 & MLP  &  xyz & 1024  & 93.1  \\
			DensePoint~\citep{liu2019densepoint} & ICCV'19 & MLP  &  xyz & 1024  & 93.2  \\
			FPConv~\citep{lin2020fpconv}  & CVPR'20  & Conv.  &  xyz& 1024     & 92.5 \\
			PointASNL~\citep{yan2020pointasnl}& CVPR'20 & Conv. &  xyz& 1024     & 92.9  \\
			DenX-Conv~\cite{lee2021connectivity} & PR'21 & Conv.   &  xyz & 1024 & 92.5   \\
			PointTrans.~\cite{zhao2021point} & ICCV'21 & Trans. &  xyz&1024  & 93.7 \\
			PAConv~\citep{xu2021paconv} w/o vot.& CVPR'21& Conv. &   xyz&1024  & 93.6 \\
			PAConv~\citep{xu2021paconv} w/ vot. & CVPR'21& Conv. &   xyz&1024  & 93.9  \Bstrut\\
			PatchFormer~\cite{zhang2022patchformer} & CVPR'22 & Trans. &  xyz&1024  & 93.5 \\
			\textbf{MKConv} w/o vot. & - & Conv.  &   xyz&1024  & \textbf{93.7} \\
			\textbf{MKConv} w/ vot.  & - & Conv.  &   xyz&1024   & \textbf{94.0}  \\
			\hline
		\end{tabular}
		\label{tab:cls}%
	\end{center}
\end{table*}

We compare the overall accuracy (OA) for the proposed MKConv with relevant previous state-of-the-art models in Table~\ref{tab:cls}. The operations of methods are categorized as point convolution, pointwise MLP and transformer. The difference between operations of convolution and MLP is described in Section~\ref{related}. We present the performance with and without the voting strategy, as in \cite{liu2019relation,xu2021paconv}, wherein an object’s class is repeatedly predicted using random scaling and the results are averaged. MKConv achieves the highest accuracy among considered models by achieving 93.7\% without voting and 94.0\% with voting.
\subsection{Object Part Segmentation}
\label{sec:psg}

We evaluate MKConv on the ShapeNetPart~\citep{yi2016scalable} dataset for object part segmentation. ShapeNetPart contains 16,881 point clouds from 16 classes. Each data is annotated with 2-6 parts, and there are 50 different parts in total. We follow the data split used in \cite{qi2017pointnet}, and 2,048 points are sampled from each point cloud. The one-hot encoding of the object label is concatenated to the last layer as in \cite{qi2017pointnet}. We adopt the same augmentation strategy used in the object classification task. 

\begin{table}[t]
	\scriptsize
	\begin{center}
		\caption{Object part segmentation results (\%) on ShapeNetPart. ``nr'', ``vot.'', ``Conv.'', ``Trans.'' denote normal vectors, voting strategy, convolution, and transformer, respectively.}
		\footnotesize
		\begin{tabular}{l|cccc|c}
			\toprule
			Method & Reference & Operation & Input & \#points & mIoU \\
			\midrule
			\midrule
			PointNet++ \citep{qi2017pointnet++}& NIPS'17 & MLP & xyz, nr & 2048	& 85.1	\Tstrut\\
			SO-Net  \citep{li2018so} & CVPR'18 & MLP		& xyz, nr & 1024	&84.6	\\
			SpiderCNN \citep{xu2018spidercnn} & ECCV'18 &Conv.& xyz, nr & 2048&85.3	\\
			PointConv \citep{wu2019pointconv} & CVPR'19 &Conv.& xyz, nr & 2048&85.7	\\
			KPConv \citep{thomas2019kpconv} & ICCV'19 &Conv.	& xyz & 2300& 86.4	\\
			CRFConv~\cite{yang2022continuous} & PR'22 & Conv. 	& xyz, nr & - & 85.5	\Bstrut\\
			\midrule
			PointNet~\citep{qi2017pointnet}& CVPR'17 &MLP & xyz & 2048	& 83.7		\\
			PointCNN \citep{li2018pointcnn} & NIPS'18 & Conv.	& xyz & 2048& 86.1	\\
			DGCNN \citep{wang2019dynamic} & TOG'19	& MLP	& xyz& 2048 &85.1	\\
			DensePoint~\citep{liu2019densepoint}& ICCV'19 & MLP   &  xyz & 2048  & 86.4 \\
			InterpCNN \citep{mao2019interpolated} & ICCV'19 & Conv. & xyz& 2048	& 86.3	\\
			RS-CNN \citep{liu2019relation} w/o vot. & CVPR'19 & Conv. & xyz & 2048 &85.8\\
			RS-CNN \citep{liu2019relation} w/ vot. & CVPR'19 & Conv. & xyz & 2048 &86.2\\
			PointASNL \citep{yan2020pointasnl} & CVPR'20 & Conv.  & xyz & 2048	& 86.1  \\
			DenX-Conv~\cite{lee2021connectivity} & PR'21 & Conv. & xyz& 2048	& 86.5	\\
			PointTrans.~\cite{zhao2021point} &  ICCV'21 & Trans. & xyz& 2048	& 86.6	\\
			PAConv \citep{xu2021paconv} w/o vot. & CVPR'21  & Conv.	& xyz& 2048	& 86.0	\\
			PAConv \citep{xu2021paconv} w/ vot. & CVPR'21 & Conv. 	& xyz& 2048	& 86.1	\\
			PatchFormer ~\cite{zhang2022patchformer}  & CVPR'22 & Trans. 	& xyz& 2048	& 86.5	\\
			\textbf{MKConv} w/o vot. & - & Conv.	& xyz & 2048	& \textbf{86.5}		\\
			\textbf{MKConv} w/ vot. & - & Conv.	& xyz & 2048	& \textbf{86.7}		\\
			\hline
		\end{tabular}
		\label{tab:psg}%
	\end{center}
\end{table}

Table~\ref{tab:psg} summarizes the part segmentation results in terms of the mean of the instance-wise intersection over union (mIoU). The operations of methods are categorized as in the classification task. Without voting, MKConv outperforms all convolution-based methods with an mIoU of 86.5\%. Examples of object part segmentation results are visualized in Figure~\ref{fig:shapenet}, verifying that MKConv performs robustly on diverse objects. More examples can be found in the supplementary material.

\begin{figure}[t]
	\centering
	\includegraphics[width=0.8\linewidth]{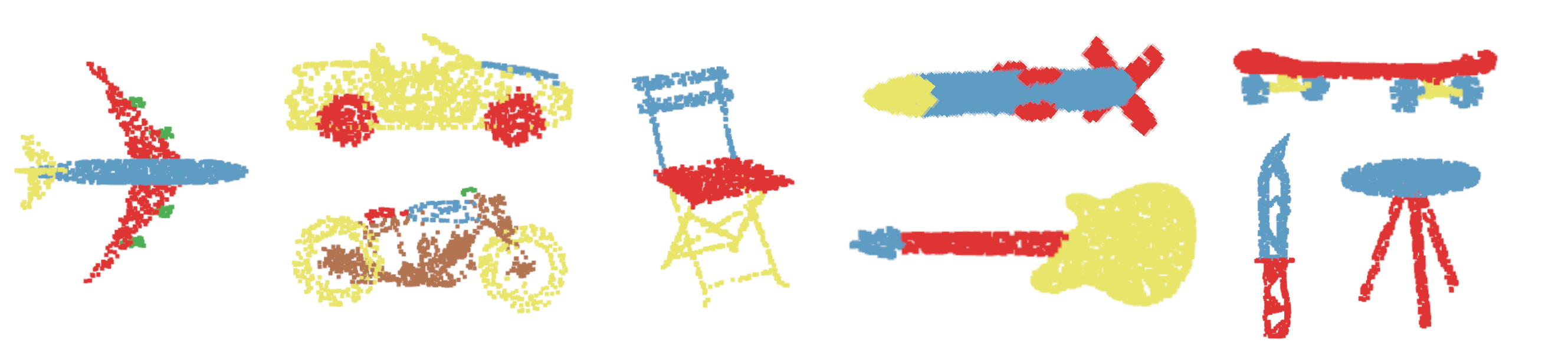}
	\caption{Visualization of part segmentation results on ShapeNetPart.}
	\label{fig:shapenet}
\end{figure} 
\subsection{Scene Semantic Segmentation}
\label{sec:sceneseg}
Unlike synthetic datasets used in classification and part segmentation, datasets for scene segmentation are generally sourced from the real-world, making the task challenging. We use S3DIS~\citep{armeni20163d}, which contains point clouds from six large-scale indoor areas. Each point is annotated with one semantic label from 13 classes. We follow the sampling strategy used in \cite{lin2020fpconv} to prepare the training data. Each point is represented by a 9D vector combining XYZ, RGB, and the normalized location. 

\begin{figure}[!t]
	\begin{center}
		\includegraphics[width=1\linewidth]{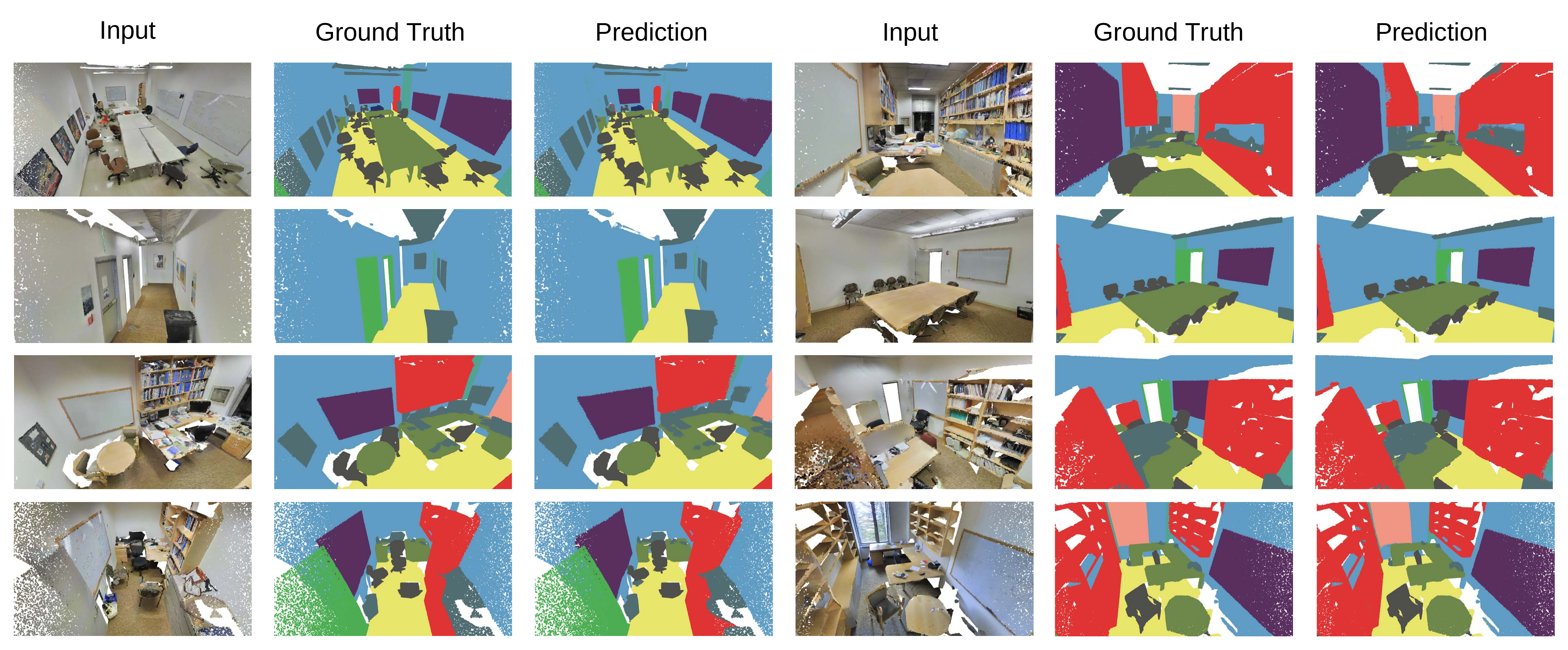}
	\end{center}
	\vspace{-0.7cm}
	\caption{Visualization of semantic segmentation results on S3DIS Area-5.
	}
	\label{fig:s3dis}
\end{figure}

For evaluation, we report the results of testing on Area-5 while training on the rest. We use evaluation metrics including the overall point-wise accuracy (OA), mean of class-wise accuracy (mAcc), and mean of the class-wise intersection over union (mIoU). As presented in Table~\ref{tab:s3dis}, MKConv shows superior performance on all metrics by achieving an OA of 89.6\%, mAcc of 75.1\%, and mIoU of 67.7\%. The qualitative results are shown in Figure~\ref{fig:s3dis}. 
\begin{table}[t]
	\caption{Semantic segmentation results evaluated on S3DIS Area-5.}
	\vspace{-0.8cm}
	\begin{center}
		\footnotesize
		\resizebox{\linewidth}{!}{
			\begin{tabular}{l | c | c | c | *{13}{c}}
				\toprule
				Method	 & OA & mAcc & mIoU & ceil.	 & floor	 & wall	 & beam	 & col.	 & wind.	 & door	 & chair	 & table	 & book.	 & sofa	 & board & clut.	\\
				\midrule
				\midrule
				PointNet \citep{qi2017pointnet} & - & 49.0	& 41.1	& 88.8	& 97.3	& 69.8	& 0.1	& 3.9	& 46.3	& 10.8	& 52.6	& 58.9	& 40.3	& 5.9	& 26.4	& 33.2	\\
				SegCloud \citep{tchapmi2017segcloud} & - & 57.4	& 48.9	& 90.1	& 96.1	& 69.9	& 0.0	& 18.4	& 38.4	& 23.1	& 75.9	& 70.4	& 58.4	& 40.9& 13.0	& 41.6	\\
				PointCNN \citep{li2018pointcnn}	&85.9& 63.9 &57.3& 92.3& 98.2& 79.4& 0.0 &17.6& 22.8& 62.1&80.6 &74.4 &66.7&31.7 &62.1& 56.7	\\
				PointWeb \citep{zhao2019pointweb}&87.0& 66.6& 60.3 &92.0 &98.5 &79.4& 0.0 &21.1 &59.7 &34.8&88.3 &76.3 & 69.3 &46.9&64.9 &52.5	\\
				PointASNL \citep{yan2020pointasnl}&87.7  &68.5 &62.6 &94.3 &98.4 &79.1 &0.0& 26.7 &55.2&66.2& 86.8 & 83.3 &68.3&47.6& 56.4& 52.1	\\
				FPConv~\citep{lin2020fpconv}	&88.3  &68.9 &62.8 &94.6 &98.5 &80.9 &0.0& 19.1 &60.1&48.9& 88.0 & 80.6 &68.4&53.2& 68.2& 54.9	\\
				SegGCN \citep{lei2020seggcn}  & 88.2 & 70.4	& 63.6	& 93.7	& 98.6	& 80.6	& 0.0	& 28.5	& 42.6&88.7	& 74.5	&  71.3 & 80.9& 69.0	& 44.4	& 54.3 \\
				CRFConv~\cite{yang2022continuous} & 89.2 &73.7 &66.2& 93.3 &96.3 &82.2 &0.0 &23.7 &60.3 &68.2& 86.0& 82.4 &73.8& 63.4& 72.4& 58.9 \\
				PAConv \cite{xu2021paconv}  & - & 73.0	& 66.6	& 94.6	& 98.6	& 82.4	& 0.0	& 26.4	& 58.0	& 60.0	& 89.7	&80.4	& 74.3	& 69.8	& 73.5	& 57.7 \\
				KPConv \citep{thomas2019kpconv}  & - & 72.8	& 67.1	& 92.8	& 97.3	& 82.4	& 0.0	& 23.9	& 58.0	& 69.0	& 91.0	&81.5	& 75.3	& 75.4	& 66.7	& 58.9 \\
				\midrule
				\textbf{MKConv} 	& \textbf{89.6}	& \textbf{75.1}& \textbf{67.7}	& 92.4	&98.2	& 83.9	& 0.0	& 28.5	& 64.5	& 65.7	& 89.7	& 82.4	& 73.9	& 67.5	& 77.3	& 55.9\\
				\hline
		\end{tabular}}
	\end{center}
\vspace{-0.6cm}
	\label{tab:s3dis} 
\end{table}

\subsection{Performance Analysis}
\noindent\textbf{Object-level tasks.}\hspace{0.3cm}In Section 4.2.1 and Section 4.2.2, we report the performance of relevant state-of-the-art models with various operations including point convolution, pointwise MLP and transformer. Notably, our method outperforms all point convolution models due to its unique multidimensional kernel. Unlike standard 1D feature representations which have only channel correlation, our multidimensional features incorporate spatial correlation in addition to channel correlation, representing the embedded information with more flexible relationships among features. Furthermore, our model demonstrates superior performance over strong transformer-based models, which are generally considered state-of-the-art in point cloud processing. These results prove that our spatially expanded feature representation shows effectiveness in capturing detailed local geometry of object-level point clouds.

\begin{figure}[t]
	\begin{center}
		\includegraphics[width=1\linewidth]{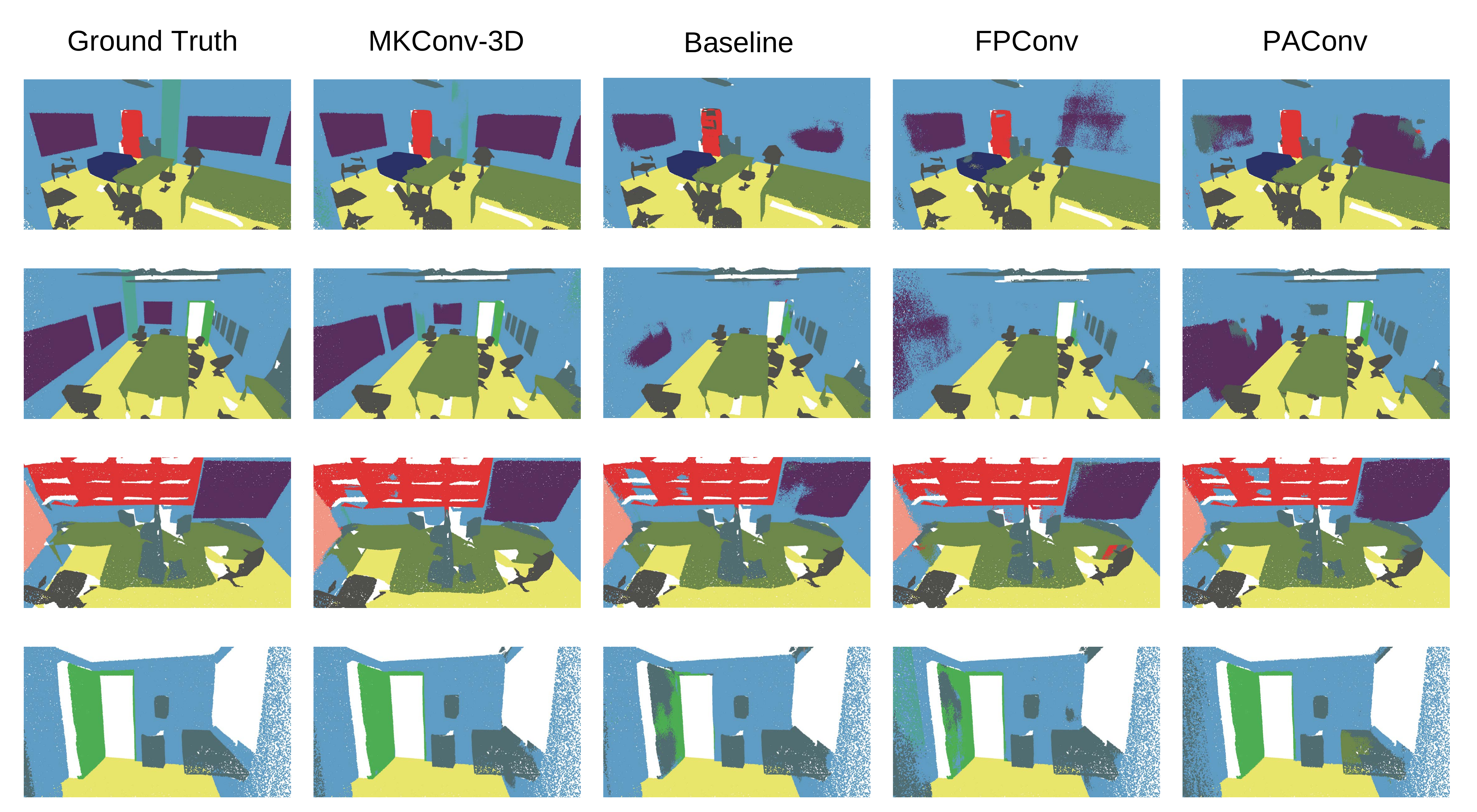}
	\end{center}
	\vspace{-0.6cm}
	\caption{Qualitative comparison on S3DIS Area-5.
	}
	\label{fig:s3dis_comp}
	\vspace{-0.6cm}
\end{figure}

\medskip
\noindent\textbf{Scene-level task.}\hspace{0.3cm}In Table~\ref{tab:s3dis}, MKConv shows a superior ability to understand the complex geometry in scene-level point clouds. The visualization shown in Figure~9 also demonstrates that MKConv clearly distinguishes between various objects that are densely located. Furthermore, in Figure~\ref{fig:s3dis_comp}, we provide a qualitative comparison of MKConv with the baseline model, PAConv~\cite{xu2021paconv}, and FPConv~\cite{lin2020fpconv}, which are the most related works to ours. The comparison with the baseline model is described in Section~\ref{sec:ablation}. Our proposed MKConv outperforms related works by effectively capturing intricate details of each instance in complex environments. While other methods struggle to differentiate between closely adjacent objects, such as the board (depicted in purple) and the wall (depicted in sky-blue), MKConv successfully separates those objects, resulting in clear and accurate semantic segmentation. This demonstrates the superiority of our method and underscores its potential for real-world applications. We can infer that the capability of MKConv comes from its multidimensional feature representation, which can better represent the local structure with additional spatial feature dimensions.

These experimental results highlight the robustness and versatility of MKConv as a general and basic feature extractor for various tasks. As our experiments are based on simple network architectures, we believe that our method can be further improved by using task-specific strategies or networks.

\subsection{Ablation Study}
\label{sec:ablation}

To evaluate the influence of various components of MKConv, we further conduct ablation studies on object classification, object part segmentation, and scene semantic segmentation. All results for object-level tasks reported in this section are obtained without the voting scheme.

\begin{table*}[!t]
	\begin{center}
		\caption{Ablation study on feature representation. Feature dimension involves one channel dimension and extra spatial dimensions. Dimensions for each point and results w/o MLA are shown.}
		\vspace{-0.2cm}
		\resizebox{\linewidth}{!}{
			\begin{tabular}{c|c|c|c|c|ccc}
				\toprule
				\multirow{2}{*}{Model} & Kernel Weight  & Feature& Spatial & Aggregation& ModelNet40 & ShapeNetPart & S3DIS \\ 
				& Dimension $k$ & Dimension & Dimension & Layer & OA & mIoU & mIoU \\
				\midrule
				\midrule
				Baseline & 1D & 1D(flattened) & \ding{55} &1x1 convolutions & 92.6 &85.7 & 63.9 \\
				MKConv-1D & 1D & 2D & 1D (64) & 1D convolutions &  93.0 & 85.8 & 65.2 \\
				MKConv-2D & 2D & 3D & 2D (8$\times$8) & 2D convolutions &  93.1 & 86.1 & 65.6 \\
				MKConv-3D & 3D & 4D & 3D (4$\times$4$\times$4) & 3D convolutions &  \textbf{93.3} & \textbf{86.3} &  \textbf{67.0} \\
				\bottomrule
		\end{tabular}}
		
		\label{tab:spatial}
	\end{center}
\end{table*}

\medskip
\noindent\textbf{Spatial correlation within multidimensional features $f_m^L$.}\hspace{0.3cm}Compared with standard feature representation which does not have a spatial dimension, the multidimensional features $f_m^L$ can contain detailed local information based on their spatial arrangements in feature space. We validate this theoretical analysis by conducting an ablation study on feature representation in Table~\ref{tab:spatial}. We test our methods with various multidimensional feature representations by adopting different dimensions for kernel weights and discrete convolutions. To verify the direct influence of feature representation, none of the models contain MLA, and all of them exploit the outer product between kernel weights and a feature vector. For a fair comparison, we seek the best performance of all models while constraining the model size to be similar. Unlike our models, the baseline adopts scalar feature representation like standard point convolution methods where kernel weights do not have spatial feature dimensions. The 2D output of the outer product in the baseline is flattened to a 1D feature vector, taking 1D feature representation comprised of channels. This process is almost identical to the memory-efficient version of PointConv~\citep{wu2019pointconv}. While the feature in baseline model represents its information only using the correlation within channel dimension, our models take advantage of spatial correlation as well as channel correlation, achieving higher performance in all tasks. The models that incorporate more spatial feature dimensions demonstrate greater performance improvement, with the MKConv utilizing 3D spatial dimensions achieving the most optimal results. As $k \geq \text{4}$ involves a heavy computational cost of k-dimensional convolutions on the large feature size, we leave it as a future work. The most remarkable performance of multidimensional feature representations is from scene semantic segmentation on S3DIS, which requires more detailed understanding of geometric traits than object-level tasks. Furthermore, the qualitative comparison of MKConv-3D with the baseline model in Figure~\ref{fig:s3dis_comp} demonstrates a substantial improvement in the ability to understand the complex geometry. Most misclassified points in the baseline model are corrected by MKConv. \begin{figure}[t]
\begin{center}
	\includegraphics[width=0.8\linewidth]{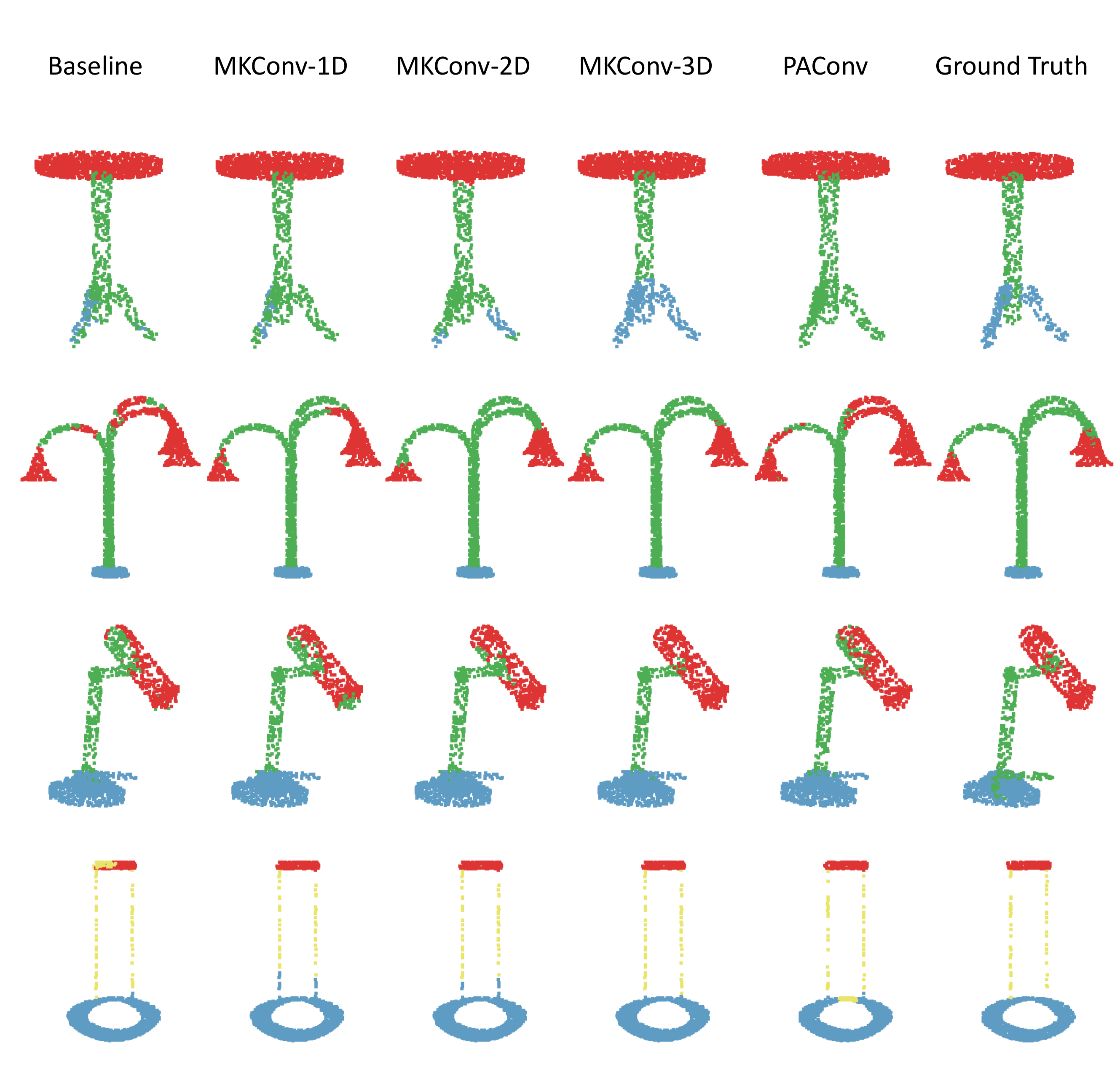}
\end{center}
\vspace{-0.7cm}
\caption{Qualitative comparison on ShapeNetPart.
}
\label{fig:shapenetcomp}
\vspace{-0.4cm}
\end{figure}
 In addition, the qualitative comparison of the baseline, MKConv-1D, MKConv-2D, MKConv-3D, and PAConv~\cite{xu2021paconv} on ShapeNetPart is presented in Figure~\ref{fig:shapenetcomp}. It is evident that the model with more spatial feature dimensions exhibits better performance on the part segmentation task. The MKConv-3D model shows promising results, and our second-best model, MKConv-2D, is also comparable to PAConv, demonstrating the advantages of multidimensional feature representation. We can deduce from this comparison that the performance improvement results from the activation of spatial feature dimensions induced by the interaction between dimension expansion process and discrete convolutions.

\begin{table}[t]
	\begin{center}
		\caption{Ablation study on the resolution of multidimensional features with various multidimensional kernel sizes. k=3 is adopted for MKConv.}
			\resizebox{0.65\linewidth}{!}{
		\scriptsize
		\begin{tabular}{c|cc}
			\toprule
			Multidimensional Kernel Size  & ModelNet40 & ShapeNetPart \\ 
			$v_3 \times v_3 \times v_3$ & OA & mIoU \\
			\midrule
			\midrule
			3 $\times$ 3 $\times$ 3 &    93.2 & 86.3 \\
			4 $\times$ 4 $\times$ 4  & \textbf{93.7} & \textbf{86.5}\\
			5 $\times$ 5 $\times$ 5  & 93.4 &86.4\\
			\bottomrule
		\end{tabular}}
		\label{tab:v}%
	\end{center}
\vspace{-0.7cm}
\end{table}

\medskip
\noindent\textbf{Multidimensional kernel unit size $v_k$.}\hspace{0.3cm}We explore various settings of the multidimensional kernel unit size $v_{k=3}$ for MKConv-3D in Table~\ref{tab:v}. Note that the multidimensional kernel size determines the resolution of dimension expansion, as the size of multidimensional local features $f_m^L$ is $\mathbb{R}^{\begin{scriptsize}\overbrace{v_k \times \cdots \times v_k}^k \times C_{in}\end{scriptsize}}$ (Eq.~\ref{eq:5}). 

For the experiment, weight normalization and MLA are employed for $v_3=\text{3,4,5}$. For $v_3 \geq \text{6}$, the spatial size of multidimensional features becomes 216 or more, which is unnecessarily large for representing the local point set. Given the small size of $v_3$, a single MKConv-3D operation requires only two or three 3D convolutions, alleviating the high computation generally produced in 3D convolution. The results in Table~\ref{tab:v} reveal that 4$\times$4$\times$4 is the most suitable size for the 3D kernel in object classification and object part segmentation. We can also observe that the model with $v_3=5$ is not superior to the model with $v_3=4$, which implies that multidimensional features representing the local point set need not be spatially large. Excessively high resolutions might cause inaccurate spatial mapping as the local point set does not contain a large number of points. For scene segmentation, the same kernel size $v_3=4$ is adopted, as a similar number of neighboring points $N$ is used to define the local point set.



\begin{table}[t]
	\begin{center}
		\caption{Component study of MKConv-3D on ModelNet40 and ShapeNetPart.}
		\resizebox{0.65\linewidth}{!}{
			\footnotesize
			\begin{tabular}{c|ccC{1.0cm}|cc}
				\toprule
				\multirow{2}{*}{Model} & \multirow{2}{*}{$\textsc{Norm}_{\textsc{L2}}$} & \multirow{2}{*}{$\textsc{Norm}_{\textsc{st}}$} & \multirow{2}{*}{MLA} & ModelNet40 &  ShapeNetPart  \\ 
				& & & & OA    & mIoU      \\
				\midrule
				A &   &    &    & 93.0 & 86.2\\
				B  & \ding{51}    &     &   & 93.1 & 86.3 \\
				C  &   & \ding{51}     &    &  93.3  & 86.3 \\
				D  &   &     & \ding{51}   & 93.3 &  86.4\\
				E  &  \ding{51} &     & \ding{51}   & 93.4 &  \textbf{86.5}\\
				F  &    & \ding{51}    & \ding{51}   & \textbf{93.7} & \textbf{86.5} \\
				\bottomrule
		\end{tabular}}
		\label{tab:ablation}%
	\end{center}
\end{table}

\medskip
\noindent\textbf{Multidimensional weight normalization and MLA.}\hspace{0.3cm}We verify the effectiveness of multidimensional weight normalization and MLA in Tables~\ref{tab:ablation} with $v_3=\text{4}$. 
Model A is set to learn without weight normalization and MLA. Table~\ref{tab:ablation} reveals that normalization (Models B, C) and MLA (Model D) result in slight improvement. Furthermore, we can verify the presence of a synergy between normalization and MLA, which exhibits meaningful margins (Models E, F).



\begin{figure}[t]
	\begin{center}
		\includegraphics[width=0.7\linewidth]{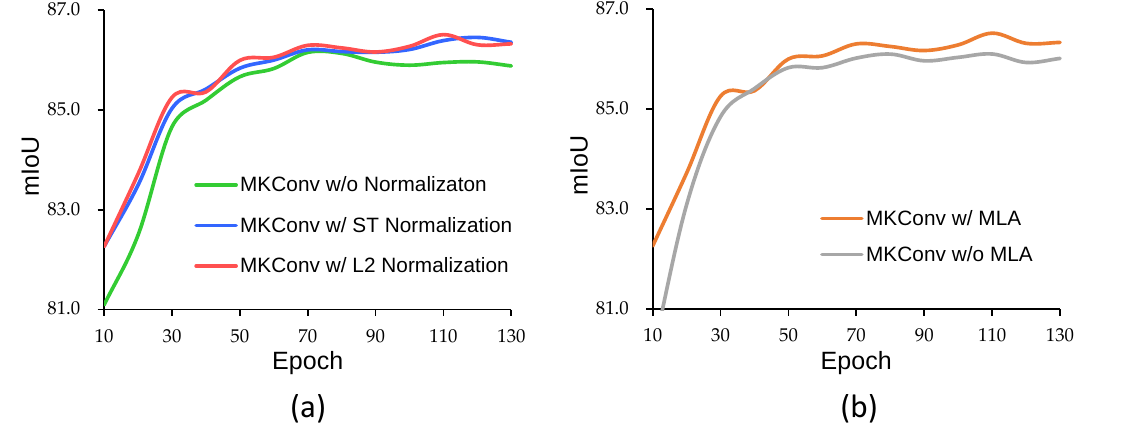}
	\end{center}
	\vspace{-0.3cm}
	\caption{Part segmentation performance (\%) on ShapeNetPart versus epochs. L2 Normalization and ST Normalization are defined in Eq.~\ref{eq:6} and Eq.~\ref{eq:7}, respectively.
	}
	\label{fig:norm}
\end{figure}

Figure~\ref{fig:norm} shows the positive influences of weight normalization and MLA during training. The multidimensional weight normalization leads to stable learning, which is especially noticeable after 70 epochs. In addition, we can see that the MLA helps achieve higher performance by reweighting the spatial feature dimensions of $f_m^L$ based on the comprehensive local structure. Although we have proposed two weight normalization methods (L2 normalization and standardization), they could be replaced with other normalization schemes.

\begin{table}[t]
	\begin{center}
		\caption{Number of parameters and latency for object classification.}
		\scriptsize
		\resizebox{0.8\linewidth}{!}{
		\begin{tabular}{c|ccccc}
			\toprule
			Method  & Input & \#points & \#params & Latency & OA\\
			\midrule 
			PointCNN~\cite{li2018pointcnn} & xyz & 1024 & 0.6M & 9.58ms  & 92.2\\
			PointConv~\citep{wu2019pointconv} & xyz, nr & 1024 &19.6M & 8.75ms  & 92.5\\
			KPConv \citep{thomas2019kpconv}& xyz & 6800 & 14.3M & 3.81ms & 92.9\\ 
			PointASNL \citep{yan2020pointasnl} & xyz & 1024 & 10.1M & 29.20ms & 93.2\\ 
			PointTrans. \cite{zhao2021point} & xyz & 1024 & 9.14M & 16.76ms & 93.7\\ 
			\textbf{MKConv-3D} & xyz & 1024 &5.63M & 6.78ms & 94.0 \\
			\bottomrule
		\end{tabular}}
		\label{tab:param}%
	\end{center}
\vspace{-0.7cm}
\end{table}

\medskip
\noindent\textbf{Model size and speed.}\hspace{0.3cm}We measure the efficiency of MKConv with a Titan RTX GPU in Table~\ref{tab:param}. Although MKConv-3D adopts spatially expanded feature representation with 3D convolutions, it shows comparable statistics to standard point convolution methods for the following reasons: 1) the dimension expansion process is independent of the number of feature channels and 2) the spatial dimensions where 3D convolutions are performed are not large as the resolution of multidimensional features $f_m^L$ is fixed to $(v_{3})^3 = 4^3$.

\section{Limitations and Future Works}

In this section, we discuss the remaining issues and provide future research directions. First, we adopted an identical multidimensional kernel size for all layers ($k$=3 and $v$=4 in case of MKConv-3D). We can further design the model that exploits a different kernel weight dimension $k$ and unit size $v_k$ for each layer. Second, we verified the influence of spatial feature dimensions in Table~\ref{tab:spatial} and Figure~\ref{fig:s3dis_comp} through empirical observation. The challenge of directly displaying the information carried by spatial dimensions in feature space remains, as discussed in Section~\ref{sec:fv}. Finally, we did not address the higher kernel weight dimensions such as $k$=4,5 in this study owing to the heavy computational cost of 4D or 5D convolutions on large feature size. The excessive dimension expansion is inefficient for representing a small local point set, but we think that expanding spatial feature dimensions while reducing the channel size can be further explored to  boost the efficiency. We believe that future work addressing these issues could make our analysis on MKConv more convincing and lead MKConv towards better feature extractor.


\section{Conclusion}
\label{con}

We presented MKConv, a novel convolution operator for point clouds that learns to expand the feature dimensions of local points using multidimensional kernels. The expanded feature representation induced by the interaction between the dimension expansion process and discrete convolutions enables the detailed geometry of local point sets to be better captured, leading to enriched feature learning. Furthermore, MLA was proposed to reweight the multidimensional local features by imparting additional spatial attention with comprehensive structure awareness. Experiments on three disparate tasks verified that our approach has superior generalization ability and performance among point convolution methods.

\title{\\\vspace{0.3cm} \normalfont Supplementary Material}

\begin{abstract}
	\begin{itemize}
		\setlength{\itemsep}{0ex}
		\item Section~\ref{sec:net} provides the network architectures, configurations and training
		parameters;
		\item Section~\ref{sec:stat} provides additional model statistics information;
		\item Section~\ref{sec:ab} presents additional ablation studies;
		\item Section~\ref{sec:vis} shows the visualization of object part segmentation and feature activation.
	\end{itemize} 
\end{abstract}

\section{Network Architectures and Training Details}
\label{sec:net}
\noindent\textbf{Network architectures.}\hspace{0.3cm}To build the networks for the classification and segmentation tasks, we adopt the hierarchical architecture of PointNet++~\citep{qi2017pointnet++} and the bottleneck residual block form of ResNet~\citep{he2016deep} as used in FPConv~\citep{lin2020fpconv} and KPConv~\citep{thomas2019kpconv}. Likewise, our network progressively processes the larger region by downsampling the point cloud with farthest point sampling (FPS) method~\citep{qi2017pointnet++}. \begin{figure}[t]
\centering
\includegraphics[width=0.8\linewidth]{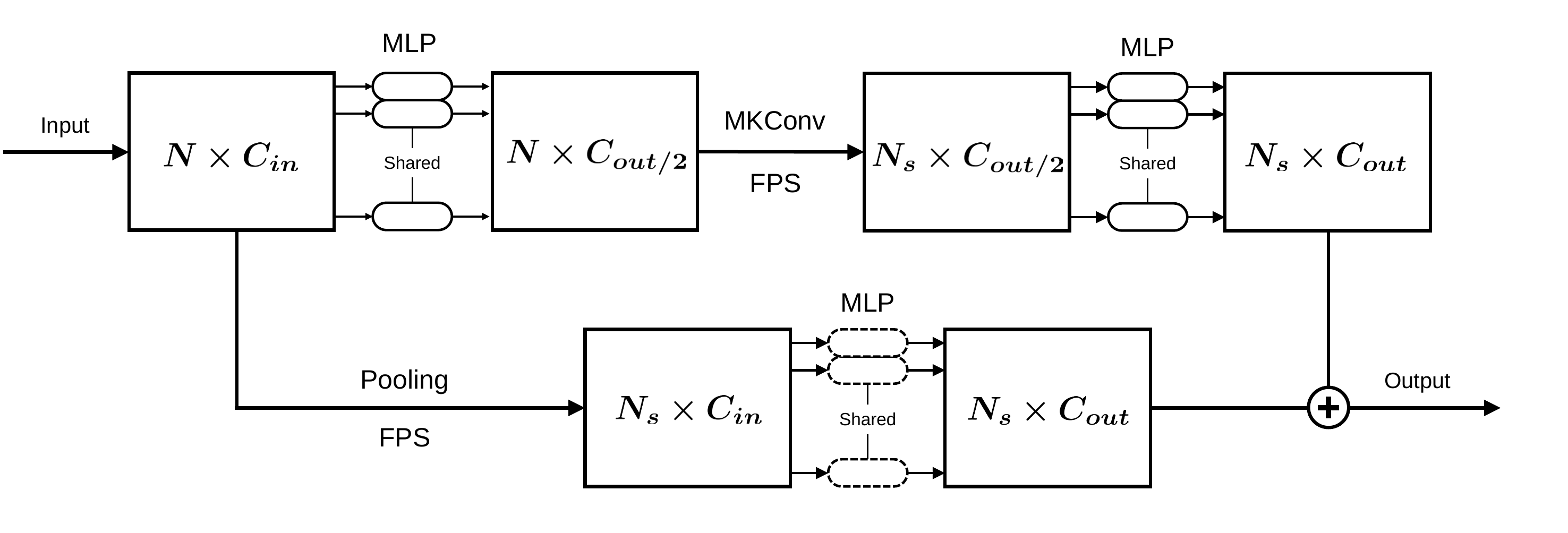}
\vspace{-0.5cm}
\caption{MKConv block with downsampling. Input is $N$ points with feature channel size $C_{in}$ and output is $N_s$ downsampled points with feature channel size $C_{out}$. Shared MLP in the shortcut is only required when $C_{in} \neq C_{out}$.}
\label{fig:block_down}
\end{figure}
\begin{figure}[t]
\centering
\includegraphics[width=0.8\linewidth]{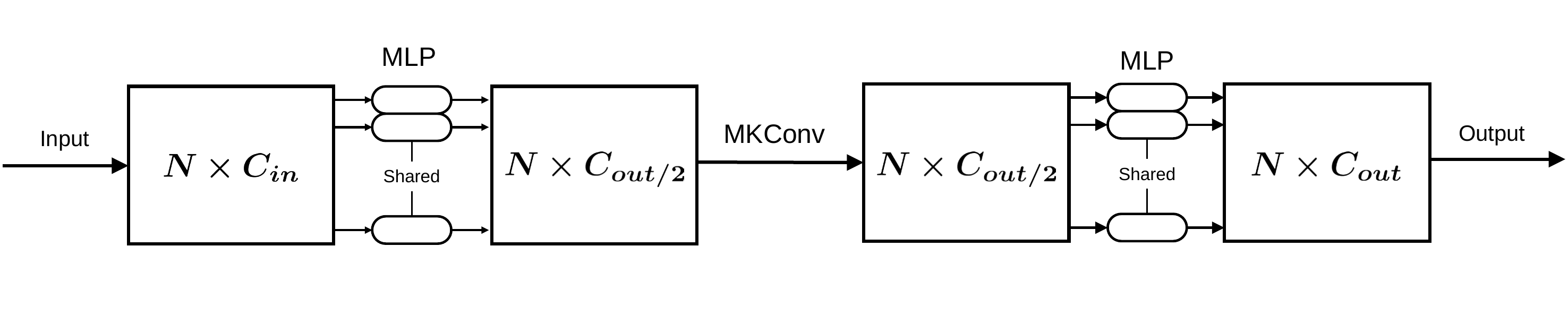}
\vspace{-0.5cm}
\caption{MKConv block without downsampling. Input is $N$ points with feature channel size $C_{in}$ and output is $N_s$ points with feature channel size $C_{out}$.}
\label{fig:block_org}
\end{figure}Each convolutional layer in the network is comprised of two MKConv blocks. The first block downsamples the point cloud using FPS and includes the shortcut as in \cite{he2016deep}, whereas the following block maintains the number of input points and does not include the shortcut. For both blocks, the shared MLP is applied before and after our MKConv, adjusting the dimension of features, which corresponds to a 1$\times$1 convolution in the ResNet block. MKConv blocks with and without downsampling are depicted in Figures~\ref{fig:block_down} and \ref{fig:block_org}, respectively. In the case of the block with downsampling, MKConv is applied on each local point set whose query point is one of the $\text{N}_\text{s}$ downsampled points. For the shortcut, max pooling operation is applied on features of points within each local point set. We apply batch normalization~\citep{ioffe2015batch} and LeakyReLU~\citep{maas2013rectifier} activation after every shared MLP and MKConv layer. 

\begin{figure}[t]
\centering
\includegraphics[width=1\linewidth]{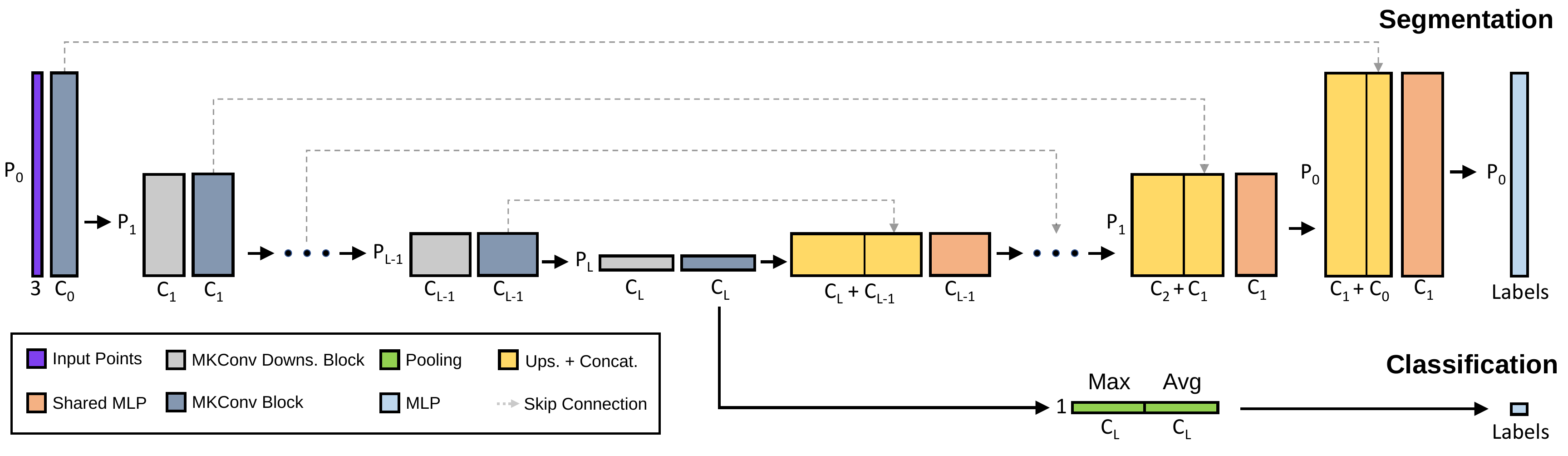}
\vspace{-0.5cm}
\caption{MKConv architectures for classification and segmentation. The output dimension of each block is indicated by P and C, where P is the number of points and C is the channel size. Network configuration for each task is provided in Table~\ref{tab:config}.}
\vspace{0.3cm}
\label{fig:net}
\end{figure}
Figure~\ref{fig:net} shows MKConv architectures for classification and segmentation. Note that two separate architectures are shown together. MKConv blocks first encode the features with multi-scale by progressively downsampling the points and increasing the radius for local point set. For classification, max pooling and average pooling are used to extract the global features, and the MLP is applied to predict an object's class. For segmentation, nearest upsampling is used to interpolate features from the previous layers, and skip connection is used to pass features from the encoder. The upsampled features and the encoder's features are concatenated, and the shared MLP fuses the concatenated features in the embedding space. Pointwise labels are then predicted with the MLP.

\begin{table}[t]
\caption{Network configurations for object classification (ModelNet40), object part segmentation (ShapeNetPart), and scene semantic segmentation (S3DIS). P is the number of points, C is the channel size, and $\text{N}$ is the number of neighboring points in local point set. $\text{r}$ and $\text{r}_\text{ds}$ are radii for local point sets in MKConv block and MKConv downsampling block, respectively.}
\vspace{-0.3cm}
\begin{center}
	\setlength\tabcolsep{2.8pt}
	\small
	\begin{tabular}{c|ccccc|ccccc|ccccc}
		\toprule
		Layer & \multicolumn{5}{c|}{ModelNet40} & \multicolumn{5}{c|}{ShapeNetPart} & \multicolumn{5}{c}{S3DIS} \\
		L  & $\text{P}$ & $\text{C}$ & N & r &$ \text{r}_\text{ds}$ & $\text{P} $& $\text{C}$ & N & r & $\text{r}_\text{ds}$& $\text{P}$ & $\text{C} $& N & r & $\text{r}_\text{ds}$\\	
		\midrule
		\midrule
		0 &1024 & 32& 20 & 0.05 &  - &2048  &64 & 32&0.1 & -& 14564&64& 32 &0.1&-\\
		1 &1024& 64 & 64 & 0.1 & 0.05 &1024 & 128 & 48&0.15 & 0.1& 8192&128&32 &0.2&0.1\\
		2 & 512 & 128 & 128 & 0.2 & 0.1 &512 & 256 & 32&0.3 & 0.15& 2048&256&32&0.4&0.2\\
		3 & 256 & 256 &32 & 0.4 &  0.2 &256 & 512 & 24&0.6 & 0.3& 512&512&32&0.8&0.4\\
		4 & 64 & 512 & 16 & 0.6 & 0.4 & 64 & 1024 & 16& 1.2 &0.6 & 128 & 1024&16&1.6&0.8\\
		\bottomrule
	\end{tabular}
	\label{tab:config}
\end{center}
\end{table}

\medskip
\noindent\textbf{Network configurations.}\hspace{0.3cm}Network configurations for object classification, object part segmentation, and scene semantic segmentation are tabulated in Table~\ref{tab:config}. The numbers of output labels are 40, 50, and 13, respectively. An ablation study on the number of neighboring points $N$ is provided in the supplementary material.

\medskip
\noindent\textbf{Training details.}\hspace{0.3cm}For object classification, we use the Momentum gradient descent optimizer with a momentum of 0.9 and an initial learning rate of 0.1. The learning rate decays with a cosine annealing until 0.001. The classification network converges in 200 epochs with a batch size of 32. 

In object part segmentation task, we use the Adam optimizer with an initial learning rate of 0.001. The learning rate decays at a rate of 0.5 every 20 epochs. The initial momentum for batch normalization is set to 0.9, and it decays at a rate of 0.5 every 20 epochs. The part segmentation network converges in 130 epochs with a batch size of 32.

For scene semantic segmentation, we use the momentum gradient descent optimizer with a momentum of 0.98, starting with a learning rate of 0.01. The learning rate decays with cosine annealing. We train our network for 200 epochs with a batch size of 8.
\section{Additional Model Statistics}
\label{sec:stat}
Table~\ref{tab:comp} presents the statistics of MKConv on various datasets and sizes of $v_3$. We provide the number of parameters and running times for training and inference. The models with multidimensional kernel unit sizes $v_3=3$ and $4$ are measured.

\begin{table}[t]
\caption{Model statistics on ModelNet40 and ShapeNetPart. The number of parameters and training/inference speeds (ms/sample) are reported.}
\vspace{-0.3cm}
\begin{center}
	\resizebox{0.8\linewidth}{!}{
		\begin{tabular}{c|cc|cc}
			\toprule
			Model & \multicolumn{2}{c|}{MKConv ($v_{3}$=3)} & \multicolumn{2}{c}{MKConv ($v_{3}$=4)} \\
			\midrule
			Dataset & ModelNet40 & ShapeNetPart & ModelNet40 & ShapeNetPart \\
			\midrule
			\midrule
			\#Points  &1024 & 2048 & 1024  & 2048\\
			\#Params  & 5.39M & 17.59M& 5.63M & 17.98M \\
			Training & 33.1ms  & 50.9ms &35.1ms & 85.0ms \\
			Inference  & 5.0ms& 7.6ms  & 6.8ms & 12.3ms \\
			\bottomrule
	\end{tabular}}
	\vspace{0cm}
	\label{tab:comp}
\end{center}
\end{table}

\section{Additional Ablation Studies}
\label{sec:ab}
\begin{figure}[t]
\centering
\includegraphics[width=0.7\linewidth]{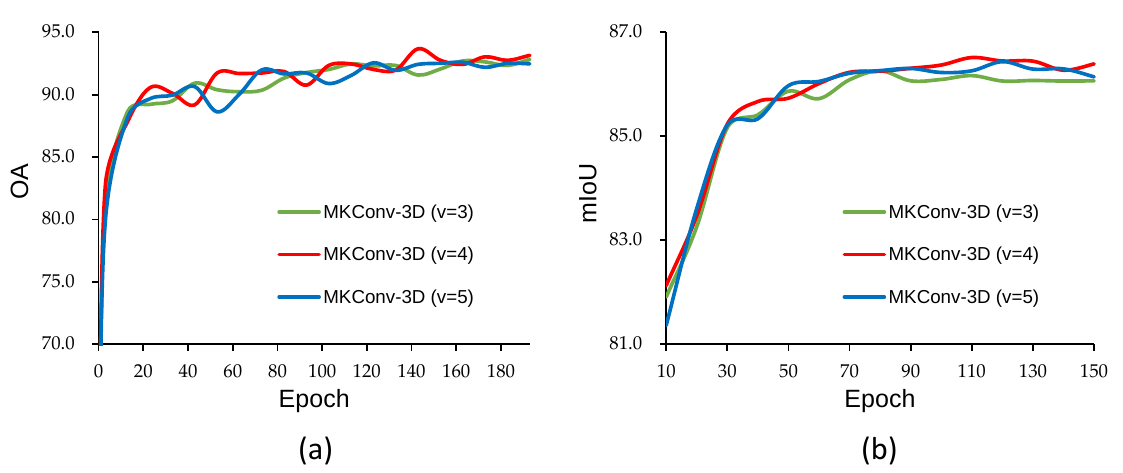}
\vspace{-0.3cm}
\caption{Object classification (a) and object part segmentation (b) performances versus epochs with different multidimensional kernel unit sizes $v_3$.}
\label{fig:v_graph}
\end{figure}

\noindent\textbf{Multidimensional kernel unit size $v$.}\hspace{0.3cm}The performance during training with various multidimensional kernel sizes is illustrated in Figure~\ref{fig:v_graph}. All kernel sizes show a similar tendency in the overall period, but $v_3=4$ achieves the best performance.

\medskip
\noindent\textbf{Ablation study on S3DIS.}\hspace{0.3cm}For scene semantic segmentation in Table~\ref{tab:ablation_s3dis}, Models B and C both achieve better performance by adopting multidimensional weight normalization schemes and MLA. In particular, MKConv with standardization and MLA (Model C) yields the best results for all evaluation metrics.

\begin{table}[t]
\caption{Component study of MKConv on S3DIS.}
\vspace{-0.3cm}
\begin{center}
	\resizebox{0.6\linewidth}{!}{
		\begin{tabular}{c|ccc|ccc}
			\toprule
			\multirow{2}{*}{Model} & \multirow{2}{*}{$\textsc{Norm}_{\textsc{L2}}$} & \multirow{2}{*}{$\textsc{Norm}_{\textsc{st}}$} & \multirow{2}{*}{MLA} & \multicolumn{3}{c}{S3DIS} \\ 
			& & & & OA & mAcc & mIoU      \\
			\midrule
			\midrule
			A &   &    &    & 89.0 & 73.5 & 66.8\\
			B  &  \ding{51} &     & \ding{51}   & 89.4 & 74.7 & 67.4 \\
			C  &    & \ding{51}  & \ding{51}   &  \textbf{89.6} &  \textbf{75.1}& \textbf{67.7}\\
			\bottomrule
	\end{tabular}}
	\label{tab:ablation_s3dis}
\end{center}
\end{table}
\medskip
\noindent\textbf{Number of neighboring points $N$.}\hspace{0.3cm}Table~\ref{tab:n} presents the results of the ablation study conducted to evaluate the impact of the number of neighboring points $N$ on object classification and segmentation tasks. Our findings suggest that the best performing model for object classification task aggregates features from a larger number of neighbors as compared to the models for segmentation tasks. This can be attributed to the fact that capturing the global structure of point clouds is crucial for the classification network, while the segmentation network needs to comprehend the detailed local geometry to predict per-point label accurately. Furthermore, our analysis shows that reducing the number of neighbors leads to a significant decrease in the inference latency. Although there is a trade-off between performance and latency, it is worthwhile to opt for a smaller number of neighbors for efficient computation since the decline in performance is minimal.

\begin{table}[t]
\begin{center}
	\caption{Ablation study on the number of neighboring points $N$ for each layer of the network, from layer 0 to layer 4.}
	\resizebox{\linewidth}{!}{
		\begin{tabular}{ccc|ccc|cc}
			\toprule
			\# of neighboring points  & ModelNet40 & Inference & \# of neighboring points  & ShapeNetPart & Inference & \# of neighboring points  & S3DIS \\ 
			for each layer & OA & Latency & for each layer  & mIoU & Latency  & for each layer & mIoU  \\
			\midrule
			\midrule
			(20, 64, 128, 32, 16) &    \textbf{93.7} & 6.76ms & (20, 64, 128, 32, 16) &    86.3 & 14.4ms & (32, 32, 32, 32, 16) &    \textbf{67.7} \\
			(20, 64, 128, 64, 32) &    93.6  & 6.99ms &(32, 48, 32, 24, 16) &    \textbf{86.5} & 12.3ms & (32, 64, 32, 32, 16) &    67.6 \\
			(20, 32, 64, 128, 32) &    93.6 & 5.87ms &(32, 64, 32, 24, 16) &    \textbf{86.5} & 12.8ms & (32, 32, 32, 16, 16) &    67.5 \\
			(32, 48, 32, 24, 16) &    93.6  & 5.57ms &(20, 20, 20, 20, 20) &    86.2 & 9.72ms & (16, 32, 64, 32, 16)&    67.5 \\
			(20, 20, 20, 20, 20) & 93.5 & 3.96ms & - &    - &- &    - & - \\
			
			\bottomrule
	\end{tabular}}
	\label{tab:n}%
\end{center}
\end{table}

\bigskip
\noindent\textbf{Number of discrete convolution layers.}\hspace{0.3cm}In Table~\ref{tab:sr}, we present an ablation study on the number of convolution layers for a single MKConv operation. The findings indicate that using only two discrete convolution layers is inadequate for effectively enforcing the learning of spatial feature dimension expansion. Moreover, the results reveal that models with three layers outperform those with two, while models with four layers achieve similar performance to those with three.
\begin{table}[H]
\caption{Ablation study on the number of convolution layers for a single MKConv operation. }
\begin{center}
	\resizebox{\linewidth}{!}{
		\begin{tabular}{c|c|cccc|ccc}
			\toprule
			\multirow{2}{*}{Model} & Spatial Resolution& \multicolumn{4}{c|}{Convolution Kernel Size}  & ModelNet40 & ShapeNetPart & S3DIS\\ 
			& of $f_{m,k}^L(p)$ & 1st kernel & 2nd kernel & 3rd kernel & 4th kernel & OA & mIoU & mIoU \\
			\midrule
			MKConv-1D & 64 &  40  & 25  & - & - &  92.8 & 85.7 & 64.8  \\
			MKConv-1D & 64 &  25  & 25  & 16  & - &  93.0 & 85.8 & 65.2  \\
			MKConv-1D & 64 &  16 & 16 & 16 & 19 &  92.9 & 85.8 & 65.2  \\
			\midrule
			MKConv-2D & 8 $\times$ 8 & 5 $\times$ 5 & 4$\times$ 4 & - & -&  93.1 & 85.9 & 65.0  \\
			MKConv-2D & 8 $\times$ 8 & 4 $\times$ 4 & 3 $\times$ 3 & 3$\times$ 3& - &  93.1 & 86.1 & 65.6  \\
			MKConv-2D & 8 $\times$ 8 & 4 $\times$ 4 & 3 $\times$ 3 & 2$\times$ 2 & 2$\times$ 2 &  93.1 & 86.1 & 65.5  \\
			\midrule
			MKConv-3D & 4 $\times$ 4 $\times$ 4 & 3 $\times$ 3 $\times$ 3 & 2 $\times$ 2 $\times$ 2  & - & - &  93.3 & 86.2 &  66.8 \\
			MKConv-3D & 4 $\times$ 4 $\times$ 4 & 2 $\times$ 2 $\times$ 2 & 2 $\times$ 2 $\times$ 2 & 2 $\times$ 2 $\times$ 2 & - &  93.3 & 86.3 &  67.0 \\
			\bottomrule
	\end{tabular}}
	\vspace{0cm}
	\label{tab:sr}
\end{center}
\end{table}

\begin{figure}[t]
\begin{center}
	\includegraphics[width=0.7\linewidth]{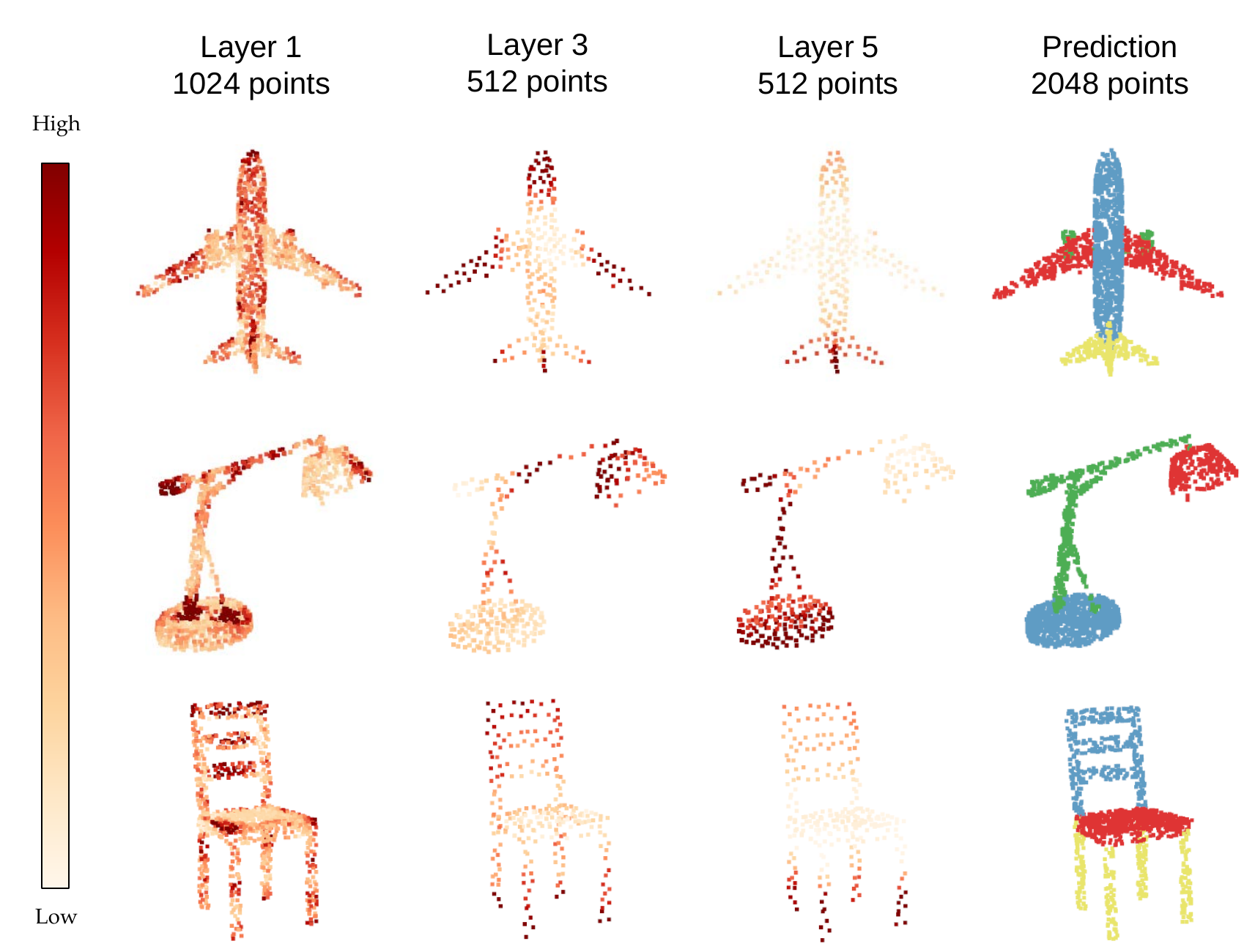}
\end{center}
\vspace{-0.1cm}
\caption{Feature activation from different layers for part segmentation on ShapeNetPart. Darker colors depict higher activation.
}
\label{fig:feature}
\end{figure}

\section{More Visualization}
\label{sec:vis}

\begin{figure}[t]
\begin{center}
	\includegraphics[width=\linewidth]{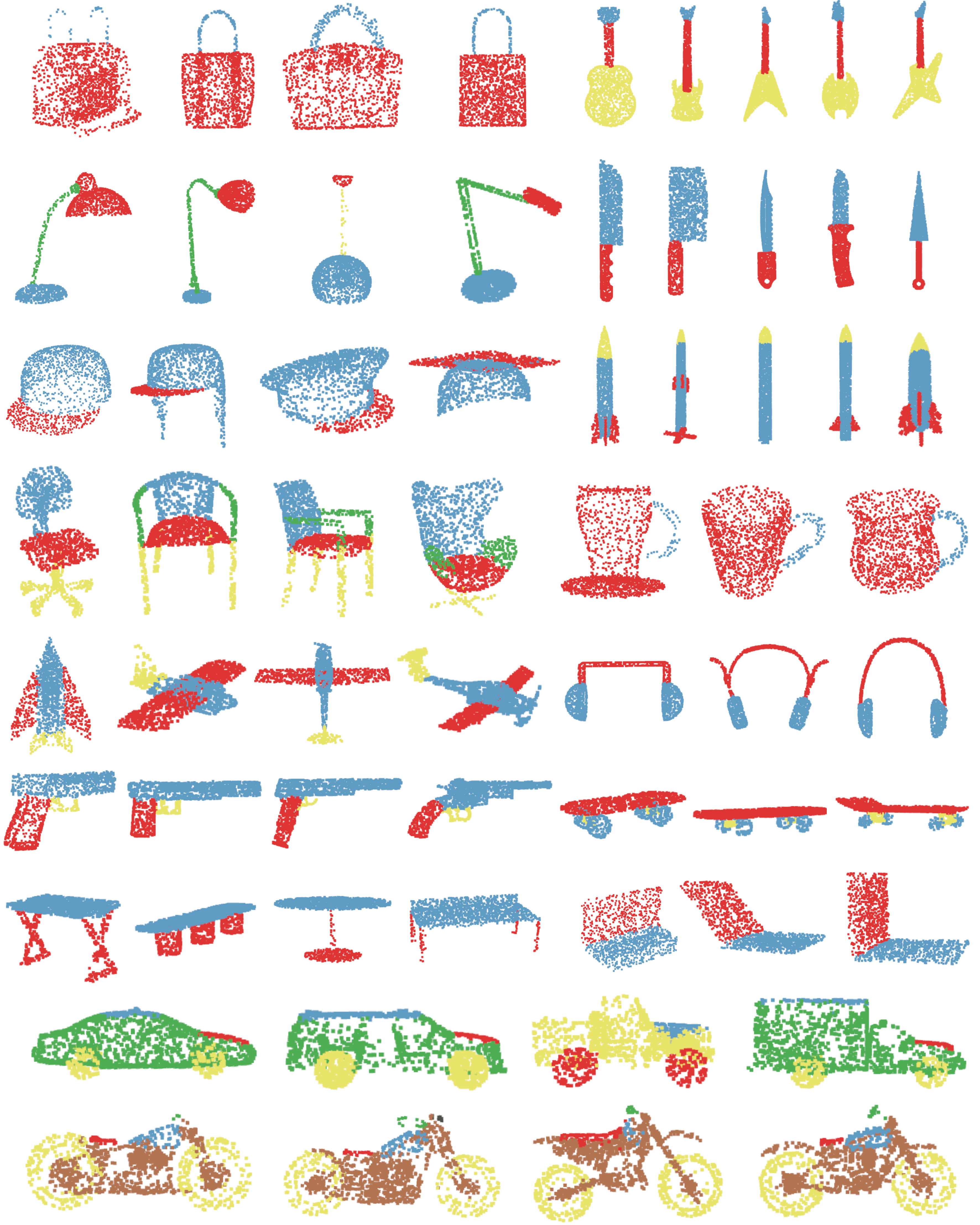}
\end{center}
\vspace{-0.1cm}
\caption{Visualization of part segmentation results on ShapeNetPart.
}
\label{fig:shapenet}
\end{figure}

\noindent\textbf{Feature visualization.}\hspace{0.3cm}Figure~\ref{fig:feature} depicts the features learned in different layers of the network. The features learned in the initial layer exhibit high activation for low-level structures such as edges and corners, while the features learned in subsequent layers exhibit high activation for semantic structures such as wings, tails, and legs. Thus, the feature response shifts from the point level to the part level by capturing more global geometric information as the layers deepen.

\medskip
\noindent\textbf{Object part segmentation.}\hspace{0.3cm}We present more visualizations of object part segmentation in Figure~\ref{fig:shapenet}.

\clearpage

\bibliography{mybibfile_clean}

\begin{thebibliography}{10}
\expandafter\ifx\csname url\endcsname\relax
  \def\url#1{\texttt{#1}}\fi
\expandafter\ifx\csname urlprefix\endcsname\relax\def\urlprefix{URL }\fi
\expandafter\ifx\csname href\endcsname\relax
  \def\href#1#2{#2} \def\path#1{#1}\fi

\bibitem{wu20153d}
Z.~Wu, S.~Song, A.~Khosla, F.~Yu, L.~Zhang, X.~Tang, J.~Xiao, 3d shapenets: A
  deep representation for volumetric shapes, in: Proceedings of the IEEE
  conference on computer vision and pattern recognition, 2015, pp. 1912--1920.

\bibitem{tchapmi2017segcloud}
L.~Tchapmi, C.~Choy, I.~Armeni, J.~Gwak, S.~Savarese, Segcloud: Semantic
  segmentation of 3d point clouds, in: 2017 international conference on 3D
  vision (3DV), IEEE, 2017, pp. 537--547.

\bibitem{graham20183d}
B.~Graham, M.~Engelcke, L.~Van Der~Maaten, 3d semantic segmentation with
  submanifold sparse convolutional networks, in: Proceedings of the IEEE
  conference on computer vision and pattern recognition, 2018, pp. 9224--9232.

\bibitem{qi2017pointnet}
C.~R. Qi, H.~Su, K.~Mo, L.~J. Guibas, Pointnet: Deep learning on point sets for
  3d classification and segmentation, in: Proceedings of the IEEE conference on
  computer vision and pattern recognition, 2017, pp. 652--660.

\bibitem{liu2019relation}
Liu, et~al., Relation-shape convolutional neural network for point cloud
  analysis, in: Proceedings of the IEEE/CVF Conference on Computer Vision and
  Pattern Recognition, 2019, pp. 8895--8904.

\bibitem{jiang2018pointsift}
M.~Jiang, Y.~Wu, T.~Zhao, Z.~Zhao, C.~Lu, Pointsift: A sift-like network module
  for 3d point cloud semantic segmentation, arXiv preprint arXiv:1807.00652
  (2018).

\bibitem{hermosilla2018monte}
P.~Hermosilla, T.~Ritschel, P.-P. V{\'a}zquez, {\`A}.~Vinacua, T.~Ropinski,
  Monte carlo convolution for learning on non-uniformly sampled point clouds,
  ACM Transactions on Graphics (TOG) 37~(6) (2018) 1--12.

\bibitem{wu2019pointconv}
W.~Wu, Z.~Qi, L.~Fuxin, Pointconv: Deep convolutional networks on 3d point
  clouds, in: Proceedings of the IEEE/CVF Conference on Computer Vision and
  Pattern Recognition, 2019, pp. 9621--9630.

\bibitem{boulch2019generalizing}
A.~Boulch, Generalizing discrete convolutions for unstructured point clouds.,
  in: 3DOR, 2019, pp. 71--78.

\bibitem{boulch2020convpoint}
A.~Boulch, Convpoint: Continuous convolutions for point cloud processing,
  Computers \& Graphics 88 (2020) 24--34.

\bibitem{thomas2019kpconv}
H.~Thomas, C.~R. Qi, J.-E. Deschaud, B.~Marcotegui, F.~Goulette, L.~J. Guibas,
  Kpconv: Flexible and deformable convolution for point clouds, in: Proceedings
  of the IEEE/CVF International Conference on Computer Vision, 2019, pp.
  6411--6420.

\bibitem{xu2021paconv}
M.~Xu, R.~Ding, H.~Zhao, X.~Qi, Paconv: Position adaptive convolution with
  dynamic kernel assembling on point clouds, in: Proceedings of the IEEE/CVF
  Conference on Computer Vision and Pattern Recognition, 2021, pp. 3173--3182.

\bibitem{yang2019learning}
Z.~Yang, L.~Wang, Learning relationships for multi-view 3d object recognition,
  in: Proceedings of the IEEE/CVF International Conference on Computer Vision,
  2019, pp. 7505--7514.

\bibitem{liu2022vfmvac}
Z.~Liu, Y.~Zhang, J.~Gao, S.~Wang, Vfmvac: View-filtering-based multi-view
  aggregating convolution for 3d shape recognition and retrieval, Pattern
  Recognition 129 (2022) 108774.

\bibitem{wu2019squeezesegv2}
B.~Wu, X.~Zhou, S.~Zhao, X.~Yue, K.~Keutzer, Squeezesegv2: Improved model
  structure and unsupervised domain adaptation for road-object segmentation
  from a lidar point cloud, in: 2019 International Conference on Robotics and
  Automation (ICRA), IEEE, 2019, pp. 4376--4382.

\bibitem{milioto2019rangenet++}
A.~Milioto, I.~Vizzo, J.~Behley, C.~Stachniss, Rangenet++: Fast and accurate
  lidar semantic segmentation, in: 2019 IEEE/RSJ International Conference on
  Intelligent Robots and Systems (IROS), IEEE, 2019, pp. 4213--4220.

\bibitem{su2018splatnet}
H.~Su, V.~Jampani, D.~Sun, S.~Maji, E.~Kalogerakis, M.-H. Yang, J.~Kautz,
  Splatnet: Sparse lattice networks for point cloud processing, in: Proceedings
  of the IEEE conference on computer vision and pattern recognition, 2018, pp.
  2530--2539.

\bibitem{qi2017pointnet++}
C.~R. Qi, L.~Yi, H.~Su, L.~J. Guibas, Pointnet++: Deep hierarchical feature
  learning on point sets in a metric space, Advances in neural information
  processing systems 30 (2017).

\bibitem{wang2019dynamic}
Y.~Wang, Y.~Sun, Z.~Liu, S.~E. Sarma, M.~M. Bronstein, J.~M. Solomon, Dynamic
  graph cnn for learning on point clouds, Acm Transactions On Graphics (tog)
  38~(5) (2019) 1--12.

\bibitem{liu2019densepoint}
Y.~Liu, B.~Fan, G.~Meng, J.~Lu, S.~Xiang, C.~Pan, Densepoint: Learning densely
  contextual representation for efficient point cloud processing, in:
  Proceedings of the IEEE/CVF International Conference on Computer Vision,
  2019, pp. 5239--5248.

\bibitem{li2018so}
J.~Li, B.~M. Chen, G.~H. Lee, So-net: Self-organizing network for point cloud
  analysis, in: Proceedings of the IEEE conference on computer vision and
  pattern recognition, 2018, pp. 9397--9406.

\bibitem{zhang2019shellnet}
Z.~Zhang, B.-S. Hua, S.-K. Yeung, Shellnet: Efficient point cloud convolutional
  neural networks using concentric shells statistics, in: Proceedings of the
  IEEE/CVF International Conference on Computer Vision, 2019, pp. 1607--1616.

\bibitem{guo2021pct}
M.-H. Guo, J.-X. Cai, Z.-N. Liu, T.-J. Mu, R.~R. Martin, S.-M. Hu, Pct: Point
  cloud transformer, Computational Visual Media 7 (2021) 187--199.

\bibitem{zhang2022pvt}
C.~Zhang, H.~Wan, X.~Shen, Z.~Wu, Pvt: Point-voxel transformer for point cloud
  learning, International Journal of Intelligent Systems 37~(12) (2022)
  11985--12008.

\bibitem{zhang2022patchformer}
C.~Zhang, H.~Wan, X.~Shen, Z.~Wu, Patchformer: An efficient point transformer
  with patch attention, in: Proceedings of the IEEE/CVF Conference on Computer
  Vision and Pattern Recognition, 2022, pp. 11799--11808.

\bibitem{lin2020fpconv}
Y.~Lin, Z.~Yan, H.~Huang, D.~Du, L.~Liu, S.~Cui, X.~Han, Fpconv: Learning local
  flattening for point convolution, in: Proceedings of the IEEE/CVF Conference
  on Computer Vision and Pattern Recognition, 2020, pp. 4293--4302.

\bibitem{klokov2017escape}
R.~Klokov, V.~Lempitsky, Escape from cells: Deep kd-networks for the
  recognition of 3d point cloud models, in: Proceedings of the IEEE
  International Conference on Computer Vision, 2017, pp. 863--872.

\bibitem{xu2018spidercnn}
Y.~Xu, T.~Fan, M.~Xu, L.~Zeng, Y.~Qiao, Spidercnn: Deep learning on point sets
  with parameterized convolutional filters, in: Proceedings of the European
  Conference on Computer Vision (ECCV), 2018, pp. 87--102.

\bibitem{li2018pointcnn}
Y.~Li, R.~Bu, M.~Sun, W.~Wu, X.~Di, B.~Chen, Pointcnn: Convolution on
  $\chi$-transformed points, in: Proceedings of the 32nd International
  Conference on Neural Information Processing Systems, 2018, pp. 828--838.

\bibitem{mao2019interpolated}
J.~Mao, X.~Wang, H.~Li, Interpolated convolutional networks for 3d point cloud
  understanding, in: Proceedings of the IEEE/CVF International Conference on
  Computer Vision, 2019, pp. 1578--1587.

\bibitem{yan2020pointasnl}
X.~Yan, C.~Zheng, Z.~Li, S.~Wang, S.~Cui, Pointasnl: Robust point clouds
  processing using nonlocal neural networks with adaptive sampling, in:
  Proceedings of the IEEE/CVF Conference on Computer Vision and Pattern
  Recognition, 2020, pp. 5589--5598.

\bibitem{lee2021connectivity}
J.~Lee, S.-U. Cheon, J.~Yang, Connectivity-based convolutional neural network
  for classifying point clouds, Pattern Recognition 112 (2021) 107708.

\bibitem{zhao2021point}
H.~Zhao, L.~Jiang, J.~Jia, P.~H. Torr, V.~Koltun, Point transformer, in:
  Proceedings of the IEEE/CVF International Conference on Computer Vision,
  2021, pp. 16259--16268.

\bibitem{yi2016scalable}
L.~Yi, V.~G. Kim, D.~Ceylan, I.-C. Shen, M.~Yan, H.~Su, C.~Lu, Q.~Huang,
  A.~Sheffer, L.~Guibas, A scalable active framework for region annotation in
  3d shape collections, ACM Transactions on Graphics (ToG) 35~(6) (2016) 1--12.

\bibitem{yang2022continuous}
F.~Yang, F.~Davoine, H.~Wang, Z.~Jin, Continuous conditional random field
  convolution for point cloud segmentation, Pattern Recognition 122 (2022)
  108357.

\bibitem{armeni20163d}
I.~Armeni, O.~Sener, A.~R. Zamir, H.~Jiang, I.~Brilakis, M.~Fischer,
  S.~Savarese, 3d semantic parsing of large-scale indoor spaces, in:
  Proceedings of the IEEE Conference on Computer Vision and Pattern
  Recognition, 2016, pp. 1534--1543.

\bibitem{zhao2019pointweb}
H.~Zhao, L.~Jiang, C.-W. Fu, J.~Jia, Pointweb: Enhancing local neighborhood
  features for point cloud processing, in: Proceedings of the IEEE/CVF
  Conference on Computer Vision and Pattern Recognition, 2019, pp. 5565--5573.

\bibitem{lei2020seggcn}
H.~Lei, N.~Akhtar, A.~Mian, Seggcn: Efficient 3d point cloud segmentation with
  fuzzy spherical kernel, in: Proceedings of the IEEE/CVF Conference on
  Computer Vision and Pattern Recognition, 2020, pp. 11611--11620.

\bibitem{he2016deep}
K.~He, X.~Zhang, S.~Ren, J.~Sun, Deep residual learning for image recognition,
  in: Proceedings of the IEEE conference on computer vision and pattern
  recognition, 2016, pp. 770--778.

\bibitem{ioffe2015batch}
S.~Ioffe, C.~Szegedy, Batch normalization: Accelerating deep network training
  by reducing internal covariate shift, in: International conference on machine
  learning, PMLR, 2015, pp. 448--456.

\bibitem{maas2013rectifier}
A.~L. Maas, A.~Y. Hannun, A.~Y. Ng, Rectifier nonlinearities improve neural
  network acoustic models, in: Proc. icml, Vol.~30, Citeseer, 2013, p.~3.

\end{thebibliography}

\end{document}